\documentclass[letterpaper]{article} %
\usepackage{microtype}
\usepackage{mathtools}
\usepackage{subfigure}
\usepackage{booktabs} %
\usepackage{times}  %
\usepackage{helvet} %
\usepackage{courier}  %
\usepackage[hyphens]{url}  %
\usepackage{graphicx} %
\usepackage{natbib}  %
\usepackage{caption} %
\usepackage{hyperref}
\usepackage{xcolor}
\usepackage{float}

\usepackage{times}
\usepackage{hyperref}
\usepackage{amsmath,amsfonts,amssymb,amsthm}
\usepackage{bm}
\usepackage{enumitem,array}
 \usepackage{algorithm, algorithmic}
\usepackage{bbm}
\usepackage{tabulary,multirow}
\usepackage{dsfont}
\usepackage{thmtools,thm-restate}
\usepackage{fullpage}

\DeclareMathOperator{\st}{s.t.}

\newcommand{\ceil}[1]{\left\lceil#1\right\rceil}
\newcommand{\floor}[1]{\left\lfloor#1\right\rfloor}

\newcommand{\Ber}{\mathrm{Ber}}

\newtheorem{lemma}{Lemma}
\newtheorem{theorem}{Theorem}
\newtheorem{definition}{Definition}

\newcommand{\R}{\mathbb{R}}

\newcommand{\NN}{{\mathbb N}}

\newcommand{\bOne}{{\bf 1}}
\newcommand{\bZero}{{\bf 0}}

\newcommand{\EE}[1]{\mathbb{E}\left[#1\right]}

\newcommand{\EEs}[2]{\mathbb{E}_{#1}\left[#2\right]}

\newcommand{\norm}[1]{\left\|#1\right\|}

\newcommand{\abs}[1]{\left|#1\right|}

\newcommand{\cD}{\mathcal{D}}

\newcommand{\cH}{\mathcal{H}}

\newcommand{\cX}{\mathcal{X}}

\newcommand{\cY}{\mathcal{Y}}

\newcommand{\bb}{{\bf b}}

\newcommand{\bC}{{\bf C}}

\newcommand{\bfm}{{\bf m}}

\newcommand{\be}{{\bf e}}

\newcommand{\bq}{{\bf q}}
\newcommand{\bu}{{\bf u}}

\newcommand{\bv}{{\bf v}}

\newcommand{\by}{{\bf y}}

\newcommand{\bx}{{\bf x}}

\newcommand{\bZ}{{\bf Z}}

\newcommand{\algfont}[1]{\texttt{#1}}

\renewcommand{\epsilon}{\varepsilon}

\renewcommand{\tilde}{\widetilde}
\renewcommand{\bar}{\overline}

\newcommand{\balpha}{{\boldsymbol \alpha}}

\newcommand{\btheta}{{\boldsymbol \theta}}

\newcommand{\bdelta}{{\boldsymbol \delta}}

\newcommand{\bmu}{{\boldsymbol \mu}}

\newcommand{\bvartheta}{{\boldsymbol \vartheta}}

\newcommand{\nothere}[1]{}

\newcommand{\err}{\mathrm{err}}

\newcommand{\opt}{\mathrm{opt}}
\newcommand{\ef}{\mathrm{ef}}
\newcommand{\eq}{\mathrm{eq}}

\usepackage{multicol}
\usepackage{authblk}

\newcommand{\bF}{\mathbf{f}}

\allowdisplaybreaks

\author[1]{Avrim Blum}
\author[2]{Nika Haghtalab}
\author[3]{Richard Lanas Phillips}
\author[1]{Han Shao}
\affil[1]{Toyota Technological Institute at Chicago, \texttt{\{avrim,han\}@ttic.edu}}
\affil[2]{University of California, Berkeley, \texttt{nika@berkeley.edu}}
\affil[3]{Cornell University, \texttt{richard@cs.cornell.edu}}

\date{}

\title{One for One, or All for All:\\Equilibria  and Optimality of Collaboration in Federated Learning}

\renewcommand{\cite}[1]{\citep{#1}}

\hypersetup{
colorlinks,
linkcolor={teal},
citecolor={teal},
urlcolor={teal}
} 
\begin{document}

\maketitle

\begin{multicols}{2}

\begin{abstract}
In recent years, federated learning has been embraced as an approach for bringing about collaboration across large populations of learning agents. However, little is known about  how collaboration protocols should take agents' incentives into account when allocating individual resources for communal learning in order to maintain such collaborations. Inspired by game theoretic notions, this paper introduces a framework for incentive-aware learning and data sharing in federated learning. Our stable and envy-free equilibria capture notions of collaboration in the presence of agents interested in meeting their learning objectives while keeping their own sample collection burden low.  For example, in an envy-free equilibrium, no agent would wish to swap their sampling burden with any other agent and in a stable equilibrium, no agent would wish to unilaterally reduce their sampling burden.

In addition to formalizing this framework, our contributions include characterizing the structural properties of such equilibria, proving when they exist, and showing how they can be computed. Furthermore, we compare the sample complexity of incentive-aware collaboration with that of optimal collaboration when one ignores agents' incentives.

\end{abstract}

\section{Introduction}

In recent years, federated learning has been embraced as an approach for enabling large numbers of learning agents to collaboratively accomplish their goals using collectively fewer resources, such as smaller data sets.
Indeed, collaborative protocols are starting to be used across networks of hospitals \citep{wen2019federated,NVIDIA2} and devices~\citep{mcmahan2017federated} and are behind important breakthroughs such as understanding the
biological mechanisms underlying schizophrenia in a large scale  collaboration of more than 100 agencies~\citep{bergen2012genome}.

This promise of creating large scale impact from mass participation has led to federated learning 
receiving substantial interest in the machine learning research community,
and has resulted in faster and more communication-efficient collaborative systems.
But, what will ultimately decide the success and impact of collaborative federated learning is the ability to recruit and retain large numbers of learning agents --- a feat that requires collaborative algorithms to
\begin{quote}
\emph{
help agents accomplish their learning objectives while ``equitably'' spreading the data contribution responsibilities among agents who want a lower sample collection burden.
}
\end{quote}
This is to avoid the following inequitable circumstances that
may otherwise arise in collaborative learning.
First, when part of an agent's data is exclusively used to accomplish another agent's learning goals; for example, if an agent's learning task can be accomplished even when she (unilaterally) lowers her data contribution.
Second, when an agent envies another agent; for example, if an agent's learning goal can be accomplished even when she swaps her contribution burden with another agent who has a lower burden.

In this paper, we introduce the first comprehensive game theoretic framework for collaborative federated learning in the presence of agents who are interested in accomplishing their learning objectives while keeping their individual sample collection burden low. 
Our framework introduces two notions of equilibria
that avoid the aforementioned inequities. 
First, analogous to the concept of Nash equilibrium~\citep{nash1951non}, our \emph{stable equilibrium} requires that no agent could unilaterally reduce her data contribution responsibility and still accomplish her learning objective. 
Second, inspired by the concept of envy-free allocations~\citep{foley1967resource,VARIAN197463},  our \emph{envy-free equilibrium} requires that no agent could swap her data contribution with an agent with lower contribution level and still accomplish her learning objective. 
In addition to capturing what is deemed as an ``equitable'' collaboration to agents, using stable and envy-free equilibria is essential for keeping learning participants fully engaged in ongoing collaborations. 

Our framework is especially useful for analyzing how the sample complexity of federated learning may be affected by the agents' desire to keep their individual sample complexities low.
To demonstrate this, we work with three classes as running examples of agent learning objectives: random discovery (aka linear) utilities, random coverage utilities, and general PAC learning utilities. Our results answer the following qualitative and quantitative questions:

\vspace*{-6pt}
\paragraph{Existence of Equilibria.}
In Section~\ref{sec:exist}, we show that the existence of a stable equilibrium depends on whether agents' learning objectives are ``well-behaved''.
In particular, we see that in the PAC learning setting, there may not exist a stable equilibrium, but under mild assumptions, a stable equilibrium exists in the random discovery (aka linear) and random coverage settings.
On the other hand, an envy-free equilibrium with equal agent contribution trivially exists. 

\vspace*{-6pt}
\paragraph{Sample Complexity of Equilibria.} In Section~\ref{sec:sample-complexity}, we show that even for well-behaved learning objectives, such as random discovery and random coverage examples, there is a large gap between the socially optimal sample complexity and the optimal sample complexity achieved by any equilibrium.
In particular, we show that there is a factor $\Omega(\sqrt{k})$ gap between the socially optimal sample complexity and that of optimal stable or envy-free equilibria for $k$ agents.

\vspace*{-6pt}
\paragraph{Algorithmic and Structural Properties.}
The main result of Section~\ref{sec:algs} shows that in the random discovery (aka linear) setting, in every
optimal stable equilibrium there is a core-set of agents for whom \emph{the equilibrium happens to also be socially optimal}, and agents who do not belong to this set make $0$ contribution in the equilibrium.
This result allows us to characterize classes of problems where the optimal stable equilibria are also socially optimal.
We further show that in some cases, linear or convex programs can be used to compute socially optimal or optimal stable equilibria.

\vspace*{-6pt}
\paragraph{Empirical Analysis.}
We show that some commonly used federated algorithms produce solutions that are very far from being an equilibrium. 
We show that the Federated-Averaging (FedAvg) algorithm of \citet{mcmahan2017communication}  lead to solutions where a large number of agents would rather reduce their contribution to as little as 25\% to 1\%.
We also work with the Multiplicative Weight Update style algorithm (MW-FED) of \citet{blum2017collaborative} and show that this algorithm produces allocations that are closer to being an equilibrium,
but more work is needed for designing algorithms  that further close this gap.

\subsection{Related Work.}

Federated learning and the model aggregation algorithm FedAvg were proposed by~\citet{mcmahan2017communication}. The collaborative learning framework of~\citet{blum2017collaborative} studied heterogeneous learning objectives in federated learning
and quantified how sample complexity improves with more collaboration.
However, except for a few recent works discussed below, agents' incentives  have not been addressed in these frameworks.

\citet{lyu2020collaborative,yu2020fairness,zhang2020hierarchically} proposed several fairness metrics for federated learning that reward high-contributing agents with higher payoffs, however, they do not consider strategicness of agents and the need for equilibrium.
\citet{li2019fair} empirically studied a different fairness notion of uniform accuracy across devices without discussing data contribution, 
while our work allows for different accuracy levels so long as every learners objective is accomplished and focuses data contribution. 
Recently, \citet{donahue2020model} studied
individual rationality in federated learning when global models may be worse than an agent's local model and used concepts from hedonic game theory to discuss coalition formation. 
Other works have discussed issues of free-riders and reputation~\citep{lin2019free,kang2019incentive} as well as markets and credit sharing in machine learning ~\citep{ghorbani2019data,agarwal2019marketplace,balkanski2017statistical,jia2019towards}.

\section{Problem Formulation}\label{sec:model-examples}

Let us start this section with a motivating example before introducing our general model in Section~\ref{sec:gen-model}.

Consider the collaborative learning problem with $k$ agents with distributions $\cD_1,\ldots,\cD_k$. For each agent $i\in[k]$, her goal is to satisfy the constraint of low expected error, i.e.,
\begin{align}
    \EEs{\{S_j\sim \cD_j^{m_j}\}_{j\in[k]}}{\err_{\cD_i}(h_{S})}\leq \epsilon\,,
    \label{eq:ind-opt}
\end{align}
for some $\epsilon>0$, where each agent $j$ takes $m_j\geq 0$ random samples $S_j\sim \cD_j^{m_j}$ and $h_S$ is a prediction rule based on $S=\cup_{j\in [k]} S_j$.
For example, $h_S$ can be the prediction output by performing ERM or gradient descent over the set $S$. Then the sample complexity of federated learning is the optimal allocation $(m_1, \dots, m_k)$ that minimizes the total number of samples conditioned on satisfying every agent's accuracy constraint,  that is,
\begin{equation}
    \begin{array}{l}
        \min \sum\limits_{i=1}^k m_i\\
        \st\,  \EEs{\{S_j\sim \cD_j^{m_j}\}_{j\in[k]}}{\err_{\cD_i}(h_{S})}\leq \epsilon,\forall i\in[k]\,.
    \end{array}
    \label{eq:social-opt}
\end{equation}
It is not surprising that optimizing Equation~\eqref{eq:social-opt} requires collectively fewer samples than the total number of samples agents need to individually solve Equation~\eqref{eq:ind-opt}\footnote{\citet{blum2017collaborative} upper and lower bound this improvement.}, but the optimal solution to Equation~\eqref{eq:social-opt} may unfairly require one or more agents  to contribute larger sample sets than could reasonably be expected from them. 
Our notion of a \emph{stable equilibrium} requires that no agent $j$, conditioned on keeping her constraint satisfied, can unilaterally reduce $m_j$. On the other hand, our \emph{envy-free equilibrium} requires that no agent $j$,
conditioned on keeping her constraint satisfied, can swap $m_j$ with another agent's contribution $m_i<m_j$.
Taking stability and envy-freeness as constraints, we ask whether such notions of equilibria always exist and whether stable and envy-free sample complexities are significantly worse than the optimal solution to Equation~\eqref{eq:social-opt}. %

\subsection{The General Framework} \label{sec:gen-model}

In this paper, we study this problem in a more general setting where there are $k$ agents and all agents collaboratively select a strategy $\btheta= (\theta_1,\ldots,\theta_k)$ from a strategy space $\Theta\subseteq \R_+^k$. Each agent $i$ selects a number $\theta_i$ as her contribution level, e.g., the number of samples.
We define $u_i:\Theta\mapsto \R$ as the utility function for each agent $i$ and her goal is to achieve 
\[
u_i(\btheta)\geq \mu_i
\]
for some $\mu_i$. This utility function is a generalization of the expected accuracy in our motivating example and $\mu_i = 1-\epsilon$ is the minimum accuracy required by the agent.

For any $\btheta\in \R^k$, $x\in \R$, let $(x,\btheta_{-i})\in\R^k$ denote the vector with the $i$-th entry being $x$ and the $j$-th entry being $\theta_j$ for $j\neq i$. 
Without loss of generality, we assume that every agent can satisfy her constraint individually, i.e., $\forall i\in[k], \exists \vartheta_i\in \R_+^k$ such that $u_i(\vartheta_i,\bZero_{-i})\geq \mu_i$ and $u_i$ is non-decreasing with $\theta_j$ for any $j$.

We say that $\btheta$ is \emph{feasible} if  $u_i(\btheta)\geq \mu_i$ for all $i\in[k]$.
We define the socially optimal solution analogously to Equation~\ref{eq:social-opt} as the optimal feasible solution that does not consider agents' incentives.

\begin{definition}[Optimal solution (OPT)]
$\btheta^\opt$ is a \emph{socially optimal solution} in $\Theta$ if it is the optimal solution to the following program
\begin{equation} \label{eq:gen-opt}
    \begin{array}{l}
        \min_{\btheta\in \Theta} \bOne^\top \btheta\\
        \st u_i(\btheta)\geq \mu_i,\forall i\in[k]\,.
    \end{array}
\end{equation}

\end{definition}

A stable equilibrium is a feasible solution where no player has incentive to unilaterally decrease her strategy.
\begin{definition}[Stable equilibrium (EQ)]
     A feasible solution $\btheta^\eq$ is a \emph{stable equilibrium} over $\Theta$ if for any $i\in [k]$, there is no $(\theta_i',\btheta^\eq_{-i})\in \Theta$ such that $\theta_i'<\theta^\eq_i$ and $u_i(\theta_i',\btheta^\eq_{-i})\geq \mu_i$.
\end{definition}

    An envy-free equilibrium is a feasible solution where no agent has an incentive to swap their sampling load with another agent.
For any $\btheta$, let $\btheta^{(i,j)}$ denote the $\btheta$ when the $i$-th and the $j$-th entries are swapped,
i.e., $\theta^{(i,j)}_i=\theta_j$, $\theta^{(i,j)}_j=\theta_i$ and $\theta^{(i,j)}_l=\theta_l$ for $l\neq i,j$. 
\begin{definition}[Envy-free equilibrium (EF)]
A feasible solution $\btheta^\ef$ is \emph{envy-free} if for any $i\in [k]$, there is no $\btheta^{\ef(i,j)}\in \Theta$ such that $\theta^\ef_j<\theta^\ef_i$ and $u_i(\btheta^{\ef(i,j)})\geq \mu_i$.

\end{definition}
We call an equilibrium $\btheta$ optimal if it is an equilibrium with minimal resources, i.e., minimizes $\bOne^\top\btheta$.

We use the game theoretic quantities known as the \emph{Price of Stability}~\citep{anshelevich2008price} and the \emph{Price of Fairness}~\citep{caragiannis2012efficiency} to quantify the impact of equilibria on the efficiency of collaboration.

\begin{definition}[Price of Stability]
Price of Stability (PoS) is defined as the ratio of the value of the optimal stable equilibrium to that of the socially optimal solution. That is, letting  $\Theta^\eq\subseteq \Theta$ be the set of all stable equilibria, $\mathrm{PoS} = \min_{\btheta\in \Theta^\eq} \bOne^\top \btheta / \bOne^\top \btheta^\opt$.
\end{definition}

\begin{definition}[Price of Fairness]
Price of Fairness (PoF) is defined as the ratio of the value of the optimal envy-free equilibrium to that of the socially optimal solution. That is, letting  $\Theta^\ef\subseteq \Theta$ be the set of all envy-free equilibria, $\mathrm{PoF} = \min_{\btheta\in \Theta^\ef} \bOne^\top \btheta / \bOne^\top \btheta^\opt$.
\end{definition}

\subsection{Canonical Examples and Settings}\label{sec:canonicalexamples}
We use the following three canonical settings as running examples throughout the paper.
\paragraph{Random Discovery aka Linear Utilities.} 
We start with a setting where any agent's utility is a linear combination of the efforts other agents put into solving the problem. As a general setting, we let $\bu(\btheta)= W\btheta$ for matrix $W\in [0,1]^{k\times k}$, where $W_{ij}$ denotes how the effort of agent $j$ affects the utility of agent $i$. We commonly assume that $W$ is a symmetric PSD matrix with an all one diagonal.

As an example, consider a setting where each agent $i$ has a distribution $\bq_i$ over the instance space $\cX$ with $\abs{\cX}=n$, and where the agent $i$ receives a reward proportional to the density of $q_{ix}$ every time an instance $x$ is realized (or discovered) by any agent's sampling effort. Formally, the utility of agent $i$ in strategy $\btheta$ is her expected reward:
\[ u_i(\btheta) = \bq_i Q^\top \btheta,
\]
where $Q=[q_{ix}]\in \R_+^{k\times n}$ denote the matrix with the $(i,x)$-th entry being $q_{ix}$, we have that $\bu(\btheta) = QQ^\top \btheta$ is a linear function.  Note that in this case, $W=  QQ^\top$ is indeed a symmetric PSD matrix.

\paragraph{Random Coverage.} 
While in our previous example an agent draws utility \emph{everytime} an instance $x$ is discovered, in many classification settings, the utility of an agent is determined by whether $x$ has been observed at all (and not the number of its observations).
This gives rise to the non-linear utilities we define below.

Consider a simple binary classification setting where the label of each point is uniformly labeled positive or negative independently of all others.
More specifically, assume that the domain $\cX$ is labeled according to a target function $f^*$ that is chosen uniformly at random from  $\{\pm 1\}^{\cX}$. Note that given any set of observed points $S = \{x_1, \dots, x_m\} \subseteq \cX$ and their corresponding revealed labels $f^*(x)$ for $x\in S$. The optimal classifier $h_S$ classifies each $x\in S$ correctly as $f^*(x)$ and misclassifies each $x\notin S$  with probability $1/2$.
Let $u_i(\btheta)$ be the expected accuracy of the optimal classifier where agent $i$ took an integral value $\theta_i$ number of samples, i.e.,
\[
u_i(\btheta) = 1 - \frac 12 \sum_{x\in \cX} q_{ix} \prod_{j=1}^k \left(1-q_{jx} \right)^{\theta_j}\,.
\]
Throughout the paper, we consider the general random coverage setting introduced here and its simpler variants where all agents' distributions are uniform over equally-sized sets.

As opposed to the linear utilities, non-integral values of $\theta_i$ (as mean of a distribution over integers)  are not as easily interpretable. Indeed, the same $\theta_i$ may refer to distributions with different expected utilities.
Here we consider one natural interpretation of a real-valued $\theta_i$: randomized rounding over $\lfloor \theta_i\rfloor$ and $\lceil \theta_i \rceil$ with mean of $\theta_i$. See Appendix~\ref{app:integral} for more information.

\paragraph{General PAC Learning.} Now we consider a general learning setting, where the labels of points are not necessarily independent. In this case, the optimal classifier can improve its accuracy on unobserved points based on those points' dependence on observed points. For example, consider a scenario where an input space $\cX$ where $\abs{\cX} =2$ and a hypothesis class that always labels points in $\cX$ either both positive or negative. Then if only one point is observed, the classifier will classify the unobserved point the same as the label of the observed one.

Generally, given input space $\cX$, hypothesis class $\cH$ and agent $i$'s distribution $\cD_i$ over $\cX$, we let utility function $u_i(\btheta)$ be the expected accuracy of any consistent function $h_S\in \cH$ given training data set $S=\cup_{j\in[k]}S_j$ when agent $i$ takes an integral value $\theta_i$ number of samples,
\[
u_i(\btheta) = 1 -{\EEs{\{S_j\sim \cD_j^{\theta_j}\}_{j\in[k]}}{\err_{\cD_i}(h_{S})}}\,.
\]
Similar to the random coverage settings, we interpret real values $\theta_i$ as the appropriate distribution over  $\lfloor \theta_i\rfloor$ and $\lceil \theta_i \rceil$ whose mean is $\theta_i$.

\section{Existence of Equilibria}\label{sec:exist}
In this section, we discuss the existence of stable and envy-free equilibria in collaborative federated learning.
Clearly, any solution with equal allocation among all agents is an envy-free allocation. That is, any feasible allocation $\btheta$ can be converted to an envy-free allocation $\btheta^\ef$ by letting $\forall i \in [k], \theta^\ef_i = \max_j \theta_j$.

\begin{theorem}\label{thm:efexist}
An envy-free solution always exists in a feasible collaborative learning problem.%
\end{theorem}
In the aforementioned envy-free solution $\btheta^\ef$, however, all agents (except for those with the maximum allocation) could unilaterally reduce their allocations while meeting their constraints, so $\btheta^\ef$ is not an equilibrium.
Indeed, in the remainder of this section we show that existence of an equilibrium in collaborative learning depends on the precise setting of the problem. In particular, we show that an equilibrium solution exists when unilateral deviations in an agent's contribution has a bounded impact on the utility of any agent.
On the other hand, an equilibrium solution may not exist if infinitesimally small changes to an agent's contribution has an outsized effect on other agents' utilities (or if an agent's strategy space is not even continuous).

We will formalize this in the next definition. Broadly, this definition states that an agent's utility increases at a positive (and bounded away from zero) rate when the agent unilaterally increases her contribution. Moreover, an agent's utility does not increase at an infinite rate. In other words, it is bounded above by a constant when other agents unilaterally increase their contributions.

\begin{definition}[Well-behaved Utility Functions]\label{assp:eqgeneral}
We say that a set of utility functions $\{u_i: \Theta \rightarrow \R \mid i\in[k]\}$ is well-behaved over $\bigtimes_{i=1}^k[0,C_i] \subseteq \Theta$ for some $C_i$s, if
and for each agent $i\in[k]$ there are constants $c^i_1 \geq 0$ and $c^i_2>0$ such that for any $\btheta\in \bigtimes_{i=1}^k[0,C_i]$,
\begin{enumerate}
    \item $\partial u_i(\btheta)/\partial \theta_i\geq c^i_2$; and
    \item for all $j\in[k]$ and $j\neq i$, $0\leq \partial u_i(\btheta)/\partial \theta_j\leq c^i_1$.
\end{enumerate} 
\end{definition}

We emphasize that the utility functions that correspond to many natural learning settings and domains, such as in the linear case and random coverage, are well-behaved.
That being said, it is also not hard to construct natural learning settings where the utility functions are not well-behaved, e.g., when an agent is restricted to taking an integral number of samples and therefore its utility is not continuous.
In the remainder of this section, we prove that, when agent utilities are well-behaved, an equilibrium exists.

\begin{theorem}\label{thm:eqexist}
For any collaborative learning problem with utility functions $u_i$s and $\mu_i$s, let $\vartheta_i$ represent the individually satisfying strategy such that $u_i(\vartheta_i,\bZero_{-i})\geq \mu_i$. If $u_i$s are well-behaved over $\bigtimes_{i=1}^k[0,\vartheta_i]$, then there exists an equilibrium.
\end{theorem}
We complement this positive result by constructing a natural learning setting that corresponds to ill-behaved utility functions and show that this problem has no equilibrium.
\begin{theorem}\label{thm:eqintexist}
There is a feasible collaborative learning problem in the general PAC learning setting that does not have an equilibrium.

\end{theorem}

\subsection{Are Canonical Examples Well-behaved?}
Recalling the three canonical examples introduced in Section~\ref{sec:canonicalexamples}, here we discuss whether they are well-behaved or not.
It is not hard to see that linear utilities are well-behaved as $u_i$ increases at a constant rate $W_{ij}$ when agent $j$ increases her strategy unilaterally, $W_{ii} = 1$, and $W_{ij}\leq 1$.

{In the random coverage case, the utilities are well-behaved over $\bigtimes_{i=1}^k[0,\vartheta_i]$ as long as 
$ u_i(\vartheta_i + 1, \bvartheta_{-i})-u_i(\bvartheta)$ is bounded away from $0$.
For example, this is the case when $\bmu \in [\frac 12, C]^k$ for $C<1$ that is bounded away from $1$.

At a high level, the smallest impact that an additional sample by agent $i$ has on $u_i$ is when $\btheta \rightarrow \bvartheta$. This impact is at least $ u_i(\vartheta_i + 1, \bvartheta_{-i})-u_i(\bvartheta) > 0$.
On the other hand, 
$\partial u_i(\btheta)/\partial \theta_j$ is bounded above, because the marginal impact of any one sample on $u_i$ is largest when no agent has yet taken a sample. Therefore, this impact is at least $u_i(1, \bZero_{-j}) - u_i(\bZero) = 1/2\sum_{x\in \cX} q_{ix} q_{jx} \leq 1/2$. 
This shows that under mild assumption the random coverage utilities are well-behaved.
}

{We note that the range of $\bigtimes_{i=1}^k[0,C_i]$ and the continuity of $\Theta$ plays an important role in determining the behavior. For example,  none of these utility functions are well-behaved over the set of integers, since $\partial u_i(\btheta)/\partial \theta_j$ is undefined.} More detail can be found in Appendix~\ref{app:wellbehave}.

\subsection{Proof of Theorem~\ref{thm:eqexist}}

In this section, we prove Theorem~\ref{thm:eqexist} and show that an equilibrium exists when utility functions are well-behaved.
Our main technical tool is to show that the best-response dynamic has a fixed point.
We define a \emph{best-response} function ${\bF:} \bigtimes_{i=1}^k[0,\vartheta_i] \mapsto \bigtimes_{i=1}^k[0,\vartheta_i]$ that maps any $\btheta$ to $\btheta'$, where $\theta'_i$ is the minimum contribution agent $i$ has to make so that $u_i(\theta'_i, \btheta_{-i})\geq \mu_i$. This is formally defined  by
$\bF(\btheta) := (f_i(\btheta))_{i\in[k]}$, where
    \begin{align*}
            f_i(\btheta) = \arg\min_{x\geq 0} u_i(x, \btheta_{-i}) \geq \mu_i\,.
    \end{align*}
Due to the monotonicity of $u_i$s and the definition of $\vartheta_i$s, it is easy to show that $f_i(\btheta)\leq \vartheta_i$.

Fixed points of function $\bF$, i.e., those $\btheta$ for which $\bF(\btheta) = \btheta$, refer to the equilibria of the collaborative learning game.
This is because, by definition, $f_i(\btheta)$ is the smallest contribution from agent $i$ that can satisfy agent $i$'s constraint in response to other agents' contributions $\btheta_{-i}$. Therefore, when $\theta_i = f_i(\btheta)$ for all $i\in [k]$, no agent can unilaterally reduce their contribution and still satisfy their constraint. That is, such $\btheta$ is an equilibrium. Therefore, to prove Theorem~\ref{thm:eqexist}, it suffices to show that the best-response function {$\bF$} has a fixed point.

\begin{restatable}{lemma}{rstfixedpt}\label{lmm:fixedpt}
If utilities are well-behaved over $\bigtimes_{i=1}^k[0,\vartheta_i]$, the best-response function $\bF$ has a fixed point, i.e., $\exists \btheta\in \bigtimes_{i=1}^k[0,\vartheta_i], \bF(\btheta) = \btheta$.
\end{restatable}

We defer the proof of Lemma~\ref{lmm:fixedpt} to Appendix~\ref{app:fixedpt}. At a high level, we show that $f$ is continuous because, for well-behaved utility functions, a small change in other agents' contributions affects the utility of agent $i$ only by a small amount. Thus, a small adjustment to agent $i$'s contribution will be sufficient to re-establish her constraint when other agents make infinitesimally small adjustments to their strategies.
Then, combining this with the celebrated Brouwer fixed-point theorem proves this lemma.

\subsection{Proof of Theorem~\ref{thm:eqintexist}}\label{sec:proofofeqexist}

In this section, we prove Theorem~\ref{thm:eqintexist} and show that an equilibrium might not exist if the utility functions are not well-behaved.
We demonstrate this using a simple example where the utility function corresponds to the accuracy of classifiers in a general PAC learning setting with integral value strategies. We give a more general construction in Appendix~\ref{app:general-consruction}.

We consider the problem in the binary classification setting where one agent's marginal distribution reveals information about the optimal classifier for another agent.

Consider the domain $\cX = \{0,\ldots,5\}$ and the label space $\cY = \{0, 1\}$. We consider agents $\{0,1,2\}$ with distributions $\cD_0,\cD_1,\cD_2$ over $\cX\times \cY$. Let $\oplus$ and $\ominus$ denote addition and subtraction modulo $3$. 

We give a probabilistic construction for  $\cD_0,\cD_1,\cD_2$. Take independent  random variables $Z_0, Z_1, Z_2$ that are each uniform over $\{0,1\}$. For each $i\in\{0,1,2\}$, distribution $\cD_i$ is a point distribution over a single instance-label pair $(2i + z_i, z_{i\ominus 1})$.
In other words, the marginal distribution of $\cD_i$ is equally likely to be the point distribution on $2i$ or $2i+1$. Moreover, the labels of points in distribution $\cD_{i\oplus 1}$ are decided according to the marginal distribution of $\cD_{i}$: If the marginal distribution of $\cD_{i}$ is a point distribution supported on $2i$ then any point in $\cD_{i\oplus 1}$ is labeled $0$, and 
if the marginal distribution of $\cD_{i}$ is a point distribution on $2i+1$ then any point in $\cD_{i\oplus 1}$ is labeled $1$.

Consider the optimal classifier conditioned on the event where agent $i$  takes a sample $(2i + z_i, z_{i\ominus 1})$ from $\cD_i$ and no other agents takes any samples. This reveals $z_i$ and $z_{i\ominus 1}$. 
Therefore, the optimal classifier conditioned on this event achieves an accuracy of $1$ for agent $i$ (by classifying $2i$ and $2i+1$ as $z_{i\ominus 1}$) 
and agent $i\oplus 1$ (by classifying $2(i\oplus 1)$ and $2(i\oplus 1)+1$ as $z_i$).
On the other hand, the optimal label for instances owned by agent $i\ominus 1$, is $Z_{i\oplus 1}$.
By the independence of random variables $Z_0, Z_1$, and $Z_2$, we have that $Z_{i\oplus 1}$ is uniformly random over $\{0,1\}$ even conditioned on $z_i$ and $z_{i\ominus 1}$. 
Therefore, the optimal classifier has an expected error of $1/2$ for agent $i\ominus 1$.
Using a similar analysis, if any two agents each take a single sample from their distributions, the accuracy of the optimal classifier for all agents is $1$.

We now formally define the strategy space and utility functions that correspond to this setting. Let $\Theta = \{0,1\}^3$ to be the set of strategies in which each agent takes zero or one sample.
Let $\bmu = \mathbf{1}$.
Let $u_i(\btheta)$ be the expected accuracy of the optimal classifier given the samples taken at random under $\btheta$.
As a consequence of the above analysis, 
$$u_i(\btheta) = \begin{cases}
    1 & \theta_i = 1 \text{ or } \theta_{i\ominus 1} = 1\\
    \frac 12 & \text{otherwise}
\end{cases}
$$

Note that any $\btheta\in \Theta$ for which $\|\btheta\|_1 \geq 2$ is a feasible solution, while no $\|\btheta\|_1 \leq 1$ is a feasible solution.
Now consider any $\btheta$ for which $\|\btheta\|_1 \geq 2$. Without loss of generality, there must be an agent $i$ such that $\theta_i = \theta_{i\ominus 1}=1$. Since $\theta_{i\ominus 1}=1$, we also have that $u_i(0, \btheta_{-i}) = 1$. That is agent $i$ can deviate from the strategy and still meet her constraint. Therefore, no feasible solution is a stable equilibrium. This proves Theorem~\ref{thm:eqintexist}.

\section{Quantitative Bounds on Price of Stability and Price of Fairness}
\label{sec:sample-complexity}

As shown in Section~\ref{sec:exist}, while an envy-free solution always exists, the existence of stable equilibria depends on the properties of the utility function.
In this section, we go beyond existence and give quantitative bounds on the sub-optimality of these equilibria notions even when they exists in the presence of (very) well-behaved functions.

\begin{restatable}{theorem}{rstflowergap}\label{thm:flowergap}
There is a collaborative learning setting with well-behaved utility functions such that the Price of Stability and Price of Fairness are at least $\Omega(\sqrt{k})$.
Moreover, these utilities correspond to two settings: a) a random domain coverage example with uniform distributions over equally sized subsets and b) a linear utility setting with $W_{ii}=1$ and $W_{ij}\in O(1/\sqrt{k})$ for $j\neq i$.
\end{restatable}

We provide an overview of the proof of Theorem~\ref{thm:flowergap} here and defer the details of this proof to Appendix~\ref{app:flowergap}.
Our construction for the random coverage and linear utility settings are very similar, here we only discuss the random coverage setting.
The crux of our approach is to build a set structure where
one agent, called the core, overlaps with all other agents and no two agent sets intersect outside of the core.
We use a relatively small $\mu_i$s so that  every agent only needs to observe one of the points in her set.
In our construction, the core is the most  ``efficient'' agent in reducing the error of all other agents and optimal collaboration puts a heavy sampling load (of about $\sqrt{k})$ on the core.
Moreover, because the core includes all the points on which two other agents intersect, the core's constraint is also easily satisfied when any other agent's constraint is satisfied. This means that in no stable or envy-free equilibrium the core can take more samples than another  agent. Therefore, most of the work has to be done by other agents in any equilibrium allocation, which requires a total of $k$ samples.
This tradeoff between being both the most ``efficient'' at sampling to reduce error and having an ``easy-to-satisfy constraint'' leads to a large Price of Stability and Price of Fairness.

\section{Structural and Algorithmic Perspectives}
\label{sec:algs}
In this section, we take a closer look at the stable equilibria of the two canonical example where they are guaranteed to exist, i.e.,  the linear utilities and the coverage utilities, and study their structural and computational aspects.

\subsection{Algorithms for Linear Utility}\label{sec:linearcvx}
Recall that linear utility functions are functions $\bu(\btheta)= W\btheta$ where $W\in [0,1]^{k\times k}$, where $W_{ij}$ denotes how the efforts of agent $j$ affects the utility of agent $i$. In this section, we assume that $W$ is a symmetric PSD matrix~\footnote{This matches our motivating use-case defined in Section~\ref{sec:model-examples}} with an all $1$ diagonal.

An immediate consequence of linear utilities is that the optimal collaborative solution can be computed using the following linear program efficiently
\begin{equation} \label{eq:LP}
\begin{array}{ll}
\min &\sum\limits_{i=1}^k \theta_i \\
\st &W\btheta \geq \bmu\\
&\btheta \geq \bZero.
\end{array}
\tag{LP 1}
\end{equation}
Interestingly, the set of stable equilibria of linear utilities are also convex and the optimal stable equilibrium can be computed using a convex program.
To see this, note that any solution to \ref{eq:LP} satisfies the constraints $\theta_i(W_i^\top \btheta -\mu_i)\geq 0,\forall i\in[k]$, where $W_i$ denotes the $i$-th column of $W$.
Hence, adding the constraints $\theta_i(W_i^\top \btheta - \mu_i)\leq 0,\forall i\in[k]$ to \ref{eq:LP} will further restrict the solution to be a stable equilibrium where $\theta_i = 0$ or $W_i^\top \btheta =\mu_i$.
Given that any stable equilibrium meets both of these constraints with tight equality of $0$, they can be equivalently represented by the following convex program.
\begin{theorem}
The following convex program computes an optimal stable equilibrium of collaborative learning with linear utility functions
\begin{equation}\label{eq:lineareq}
\begin{array}{ll}
\min &\sum\limits_{i=1}^k \theta_i \\
\st &W\btheta \geq \bmu \\
&\btheta \geq \bZero\\
&\btheta^\top W \btheta-\mu^{\top} \btheta \leq 0,
\end{array}
\tag{CP 1}
\end{equation}
where the last inequality is convex when $W$ is PSD.
\end{theorem}

\subsection{Structure of Equilibria for Linear Utility}
In this section, we take a closer look at the structural properties of stable and envy-free equilibria and provide a qualitative comparison between them and the optimal solutions.
The main result of this section is that in any optimal stable equilibrium, there is a core subset of $k$ agents for which the equilibrium is also a socially optimal collaboration, while all other agents' contributions are fixed at $0$.

\begin{restatable}{theorem}{rstthmlinearopt}\label{thm:lineareqopt}
Let $\btheta^\eq$ be an optimal stable equilibrium for linear utilities $u_i(\btheta) = W^\top_i\btheta$ and $\mu_i = \mu$ for $i\in [k]$, where $W$ is a symmetric PSD matrix.
Let $I_{\btheta^\eq} = \{ i\mid \theta_i^\eq = 0\}$ be the set of non-contributing agents and let $\bar W$ and $\bar\btheta^\eq$ be the restriction of $W$ and $\btheta^\eq$ to  $[k]\setminus I_{\btheta^\eq}$.
Then $\bar\btheta^\eq$ is \emph{a socially optimal solution} for the set of agents $i\in [k]\setminus I_{\btheta^\eq}$, i.e., agents with utilities $u_i(\bar\btheta) = \bar{W}_i^\top \bar\btheta$ for $i\in [k]\setminus I_{\btheta^\eq}$.

Furthermore, let $\tilde{\btheta}$ represent the extension of $\bar{\btheta}$ by padding $0$s at $I_{\btheta^\eq}$, i.e., $\tilde{\theta}_i=0$ for $i\in I_{\btheta^\eq}$ and $\tilde{\theta}_i = \bar{\theta}_i$ for $i\in [k]\setminus I_{\btheta^\eq}$. For any $\bar\btheta$ that is a socially optimal solution for agents $[k]\setminus I_{\btheta^\eq}$, $\tilde{\btheta}$ is an optimal stable equilibrium for agents $[k]$.
\end{restatable}

This theorem implies that any equilibrium in which all agents have non-zero contribution has to be socially optimal.

\begin{restatable}{corollary}{rstcrllinearopt}\label{cor:lineareqopt}
Consider an optimal equilibrium $\btheta^\eq$. If $\btheta^\eq > \bZero$, then $\btheta^\eq$ is socially optimal.
\end{restatable}

An advantage of Corollary~\ref{cor:lineareqopt} is that in many settings it is much simpler to verify that every agent has to contribute a non-zero amount at an equilibrium without computing the equilibrium directly. One such class of examples is when matrix $W$ is a \emph{diagonally dominant} matrix, i.e., $\sum_{j\neq i}W_{ij}< W_{ii}$ for all $i\in [k]$, in addition to satisfying the requirements of Theorem~\ref{thm:eqexist}.
In this case, every agent can satisfy their own constraint in isolation using $\vartheta_i = 1/\mu$ contribution. Therefore, in any stable equilibrium the total utility an agent will receive from all others (even at their maximum contribution of $1/\mu$) is not sufficient to meet her constraint. Therefore, 
every agent has a non-zero contribution in an equilibrium. This shows that the Price of Stability corresponding to diagonally dominant matrices is $1$. 

We defer the proofs of Theorem~\ref{thm:lineareqopt} and Corollary~\ref{cor:lineareqopt} to Appendix~\ref{app:lineareqopt}. At a high level, our proofs use the duality framework and the linear program~\eqref{eq:LP} and convex program~\eqref{eq:lineareq}. 
At a high level, the first part of Theorem~\ref{thm:lineareqopt}  follows from the observation that the dual problem of the linear program~\eqref{eq:LP} for the set of agents $i\in [k]\setminus I_{\btheta^\eq}$ is
\begin{equation*}
\begin{array}{ll}
\max_\by &\bOne^\top \by \\
\st &\bar{W} \by\leq \mu \bOne\\
&\by \geq \bZero\,.
\end{array}
\end{equation*}
Since $\btheta^\eq$ is a stable equilibrium with positive entries in $\bar \btheta^\eq$, it is not hard to see that $\bar{W} \bar \btheta^\eq = \mu \bOne$. Then we know that $\bar \btheta^\eq$ is not only a feasible solution to \eqref{eq:LP} for $[k]\setminus I_{\btheta^\eq}$ but also a feasible solution to its dual with the same value. Therefore, $\bar \btheta^\eq$ is a socially optimal solution for the set of agents $i\in [k]\setminus I_{\btheta^\eq}$.
A closer look at this dual also proves that any $0$ padding of a socially optimal solution for the set of agents $[k]\setminus I_{\btheta^\eq}$ is a stable equilibrium for the set of agents $[k]$ as well.

Lastly, in the linear utilities case, it is not hard to show that any stable equilibrium is also envy-free. 

\begin{restatable}{theorem}{rstlineareqef}\label{thm:lineareqef}
When $W_{ij}< W_{ii}$ for all $i,j\in [k]$, any stable equilibrium is also envy-free.
\end{restatable}

We defer the proof of Theorem~\ref{thm:lineareqef} to Appendix~\ref{app:lineareqef}. Theorem~\ref{thm:lineareqef} and Corollary~\ref{cor:lineareqopt} together highlight an advantage of optimal stable equilibria. Not only are these equilibria are socially optimal for a subset of agents (and in some cases for all agents) but also they satisfy the additional property of being envy-free.

\subsection{Coverage Utilities}
We complement the algorithmic and structural perspective of equilibria in the linear utility case with those for the random coverage utilities. Unlike the linear utility case, both the stable feasible set and the envy-free feasible set for the random coverage utilities are non-convex, which indicates that either optimal stable equilibrium or optimal envy-free equilibrium is intractable.
\begin{restatable}{theorem}{rstthmcvrgeqnoncvx}\label{thm:cvrgeqnoncvx}
There exists a random coverage example with strategy space $\Theta=\R_+^k$ such that $\Theta^\eq$ is non-convex, where $\Theta^\eq\subseteq \Theta$ is the set of all stable equilibria. 
\end{restatable}

We defer the proof to Appendix~\ref{app:ncvxef} and provide an overview of the proof of Theorem~\ref{thm:cvrgeqnoncvx} here. Consider an example where there are $2$ agents and both are with a uniform distribution over the instance space $\cX=\{0,1\}$ and $\mu_i=3/4$ for all $i\in [2]$.
Note that both $\be_1$ and $\be_2$ are stable equilibria, since both agents receive $3/4$ utility if either of them observe any one of the instances.
Now consider a convex combination of these two strategies $(\be_1+\be_2)/2$, i.e, each agent takes one sample with probability $1/2$. In this case, there is a small probability that when both agents sample they both uncover the same point. Thus they do not receive any marginal utility from the second sample. This means that the utility that both agents receive from 
$(\be_1+\be_2)/2$ is strictly less than $3/4$, that is, $(\be_1+\be_2)/2$ is not even a feasible solution let alone a stable equilibrium. For more details refer to Appendix~\ref{app:ncvxef}.

\begin{restatable}{theorem}{rstthmcvrgefnoncvx}\label{thm:cvrgefnoncvx}
There exists a random coverage example with strategy space $\Theta=\R_+^k$ such that $\Theta^\ef$ is non-convex, where $\Theta^\ef\subseteq \Theta$ is the set of all envy-free equilibria. 
\end{restatable}

We defer the proof to Appendix~\ref{app:ncvxef}. At a high level, considering a complete graph on $4$ vertices, we let each edge correspond to one agent and put one point in the middle of every edge and one point on every vertex. Then we let each agent's distribution be a uniform distribution over $\cX_i$, which is the $3$ points on agent $i$'s edge. In this example, we can obtain a envy-free equilibrium $\btheta^\ef$ by picking any perfect matching on this complete graph and then letting $\theta_i^\ef=1$ if edge $i$ is in this matching and $\theta_i^\ef=0$ otherwise. However, we can show that there exists a convex combination of two envy-free equilibria corresponding to two different perfect matchings such that it is not envy-free.

\section{Experimental Evaluation}
To demonstrate potential issues with not considering incentives in federated learning, we compare two federated learning algorithms that account for these incentives to different extents. We consider both federated averaging \citep{mcmahan2017communication} and a collaborative PAC-inspired algorithm based on \citet{blum2017collaborative, nguyen2018improved,chen:tight2018} called MW-FED. Federated averaging is envy-free as agents take the same number of samples in expectation. Unfortunately, FedAvg may find solutions that are far from any stable equilibrium. MW-FED does not explicitly guarantee envy-freeness or stability, however, we demonstrate that it produces solutions that are closer to being a stable equilibrium. This is due to the fact that it implicitly reduces the sample burden of those agents who are close to having satisfied their constraints.

\paragraph{Federating Algorithms} At a high level, FedAvg involves sending a global model to a set of clients and requesting an updated model (from some number of updates performed by the client) based on the client’s data. The server then calculates a weighted average of these updates and sets this as the new set of parameters for the model. MW-FED uses the Multiplicative Weight Update meta-algorithm and adjusts the number of samples that each agent has to contribute over multiple rounds.
MW-FED takes a fixed number of samples at each round, but distributes the load across agents proportional to weights $w_i^t$. In the first iteration, the load is distributed uniformly between the agents, i.e., $w_i^1 = 1$. At every new  iteration, the current global model is tested on each agent's holdout set. Distributions that do not meet their accuracy objective increase their $w_i^t$ according to the Multiplicative Weight Update.
A more detailed statement of the algorithm can be found in Appendix \ref{experimental_section}.

\paragraph{EMNIST Dataset} We study the balanced split of the EMNIST \citep{DBLP:journals/corr/CohenATS17}, a character recognition dataset of 131,600 handwritten letters. EMNIST provides a variety of heterogenous data while still remaining accessible enough to run a sufficient number of trials. We encourage further heterogeneity via a sampling technique that identifies difficult and easy points. Each agent is assigned 2000 points from some mixture of these two sets.
Implicitly, this creates agents who have varying degrees of difficulty in achieving their learning objectives.
From these $2000$ points, 1600 are selected as the training and 400 as a validation set. During the course of training, we say that a distribution's contribution is the fraction of its 1600 points that it will use during learning. That is, if an agent's contribution level is $0.01$, it will take a sample of $16$ points at the beginning of the optimization procedure and only uses those data when creating mini-batches.

\begin{figure}[H]
    \centering
 \includegraphics[width=.4\textwidth]{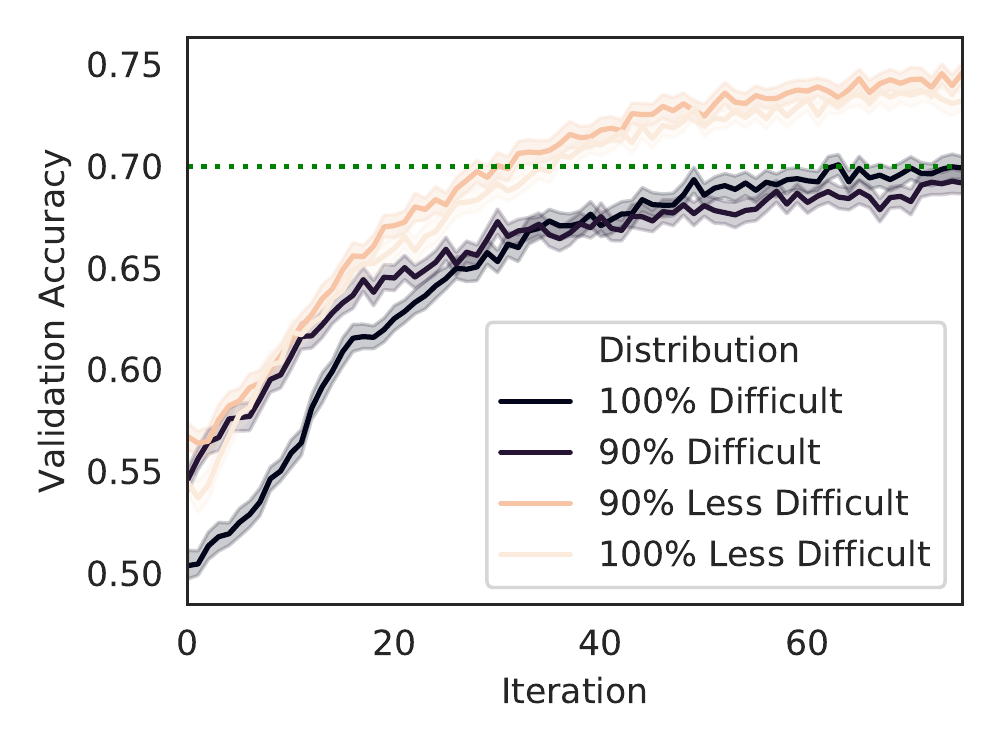}
    \caption{Each line here represents the average of 100 non-federated runs of a distribution used in this experiment. Note that the less difficult distributions reach the threshold quickly, whereas the more difficult distributions take nearly three times as long.}
    \label{fig:solo}
\end{figure}

For clarity of presentation, in these experiments we use four agents, two that have harder distributions and two that have easier distributions. Figure~\ref{fig:solo} shows the average performances of the four distributions without federation. Our observations and trends hold across larger sample sizes and with additional agents as shown in Appendix~\ref{experimental_section}. For training, we use a four-layer neural network with two convolutional layers and two fully-connected layers. For efficiency and to mirror real-world federated learning applications, we pre-train this model on an initial training set for 40 epochs to achieve 55\%  accuracy and then use federated training to achieve a 70\% accuracy level for all agents. More details on the dataset and model used can be found in Appendix~\ref{experimental_section}.

\paragraph{Results.}
To compare the two algorithms, we consider the resulting likelihood of any agent's constraint remaining satisfied when they unilaterally reduce their contribution level. Specifically, each agent wants to attain an accuracy of 70\% on their individual validation set. We chose this threshold as the easy distributions readily, individually converge above this level whereas, in our time horizon, the difficult distributions took, on average, nearly three times as long. See Figure~\ref{fig:solo} for the averaged individual performance trajectories. 
If an agent can drop their contribution level significantly and still attain this accuracy consistently during the optimization process, then either (a) other agents are oversampling and this agent is able to benefit from their over-allocation or (b) the agent was sampling too much to begin with relative to their requirements.

FedAvg makes no distinction between these cases. All agents contribute at an equal rate to convergence. On the other hand, MW-FED quickly reduces an agent's contribution level when she has met her constraints, reducing her ability to oversample.

\begin{figure}[H]
    \centering
     \includegraphics[width=.4\textwidth]{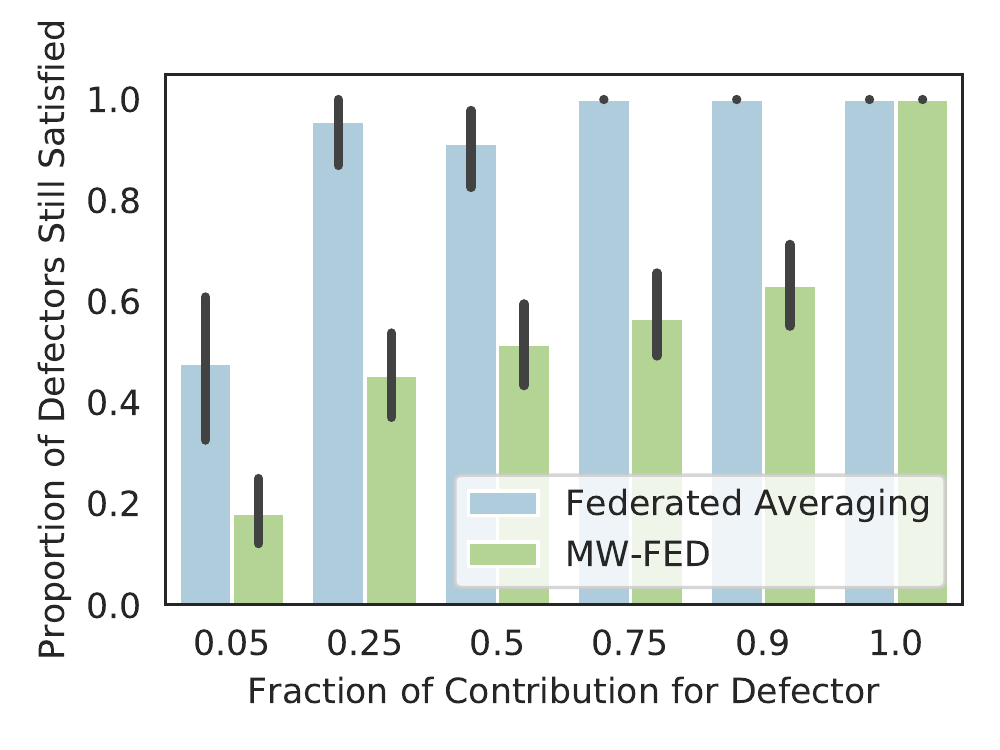}
    \caption{Comparing the likelihood that a single defector will reach their accuracy threshold at various contributions for federated averaging and MW-FED after 10 epochs. The result shows that MW-FED results in allocations that are closer to an equilibrium compared to FedAvg.
    }
    \label{fig:Doublebar}
\end{figure}

Figure~\ref{fig:Doublebar} shows the results of FedAvg and MW-FED run on the dataset 100 times. When everyone fully contributes, 100\% of these FedAvg runs satisfy the requirements of all agents by the tenth epoch. This figure compares the probability that, if a random single agent defected to a given contribution level, they would expect to have met their accuracy threshold at this point. For instance, if a single random agent only contributed 25\% of their data in FedAvg, they still have a 94\% chance of being satisfied by the tenth epoch. By comparison, only 45\% of agents at the same contribution level would succeed in MW-FED. This is striking as is discussed further in Appendix~\ref{experimental_section} where, even with pre-training, none of the agents in the individual (non-federated) setting reaches 70\% accuracy with 50\% or less of their data.
In Appendix~\ref{experimental_section}, we give one possible explanation for the performance of MW-FED by drawing parallels to algorithms in Section~\ref{sec:algs} that work in the linear setting.

\section{Conclusion}

Our paper introduced a comprehensive game theoretic framework for collaborative federated learning that considers agent incentives. Our theoretical results and empirical observations form the first steps in what we hope will be a collective push towards  designing equitable  collaboration protocols that will be essential for recruiting and retaining large numbers of participating agents.

\section{Acknowledgements}
This work was supported in part by the National Science Foundations under grants CCF-1733556 and CCF-1815011, and a J.P. Morgan Chase faculty research award. Part of this work was done while Haghtalab was visiting the Simons Institute for the Theory of Computing.

\bibliographystyle{apalike}
\bibliography{Major}

\end{multicols}

\clearpage
\appendix
\onecolumn

\section{Real-Valued Strategies} \label{app:integral}

In the examples of random coverage and general PAC learning, it is common to consider integral values of $\theta_i$. For a real-valued $\theta_i$, we consider one natural interpretation: randomized rounding over $\floor{\theta_i}$ and $\ceil{\theta_i}$. More specifically, let agent $i$ randomly draw an integral value $m_i\sim \sigma(\theta_i)$, where $\sigma(\theta_i)=\floor{\theta_i} + \Ber(\theta_i-\floor{\theta_i})$, and then uses $m_i$ as her strategy. Then the utility function is defined by taking expectation over $\bfm = (m_1,\ldots,m_k)$. That is,
\[
u_i(\btheta) = \EEs{\bfm}{1 - \frac 12 \sum_{x\in \cX} q_{ix} \prod_{j=1}^k \left(1-q_{jx} \right)^{m_j}}\,.
\]
Similarly, we define the utility function in general PAC learning as
\[
u_i(\btheta) = 1 -\EEs{\bfm}{\EEs{\{S_j\sim \cD_j^{m_j}\}_{j\in[k]}}{\err_{\cD_i}(h_{S})}}\,.
\]
Note that these definitions work for integral-valued $\theta_i$ as well.

\section{Calculation of Well-behaved Property}\label{app:wellbehave}
\paragraph{Linear Utilities.} 
The linear utilities are well-behaved over any $\bigtimes_{i=1}^k [0,C_i]\subseteq \Theta$. Agent $i$'s utility increases at a constant rate $\partial \theta_i(\btheta)/\partial \theta_i=W_{ii} = 1$ when the agent increases its strategy unilaterally and increases at rate $\partial \theta_i(\btheta)/\partial \theta_j=W_{ij}\leq 1$ when agent $j$ increases its strategy unilaterally.

\paragraph{Random Coverage.}
For any $\bigtimes_{i=1}^k [0,C_i+1]\subseteq \Theta$, if $u_i(C_i+1,\bC_{-1})-u_i(\bC)$ is bounded away from $0$ for all $i$, then the utilities are well-behaved over $\bigtimes_{i=1}^k [0,C_i]$, where $\bC =(C_1,\ldots,C_k)$. At a high level, the smallest impact that an additional sample by agent $i$ has on $u_i$ is when $\btheta \rightarrow \bC$. This impact is at least $ u_i(C_i + 1, \bC_{-i})-u_i(\bC) > 0$.
On the other hand, 
$\partial u_i(\btheta)/\partial \theta_j$ is bounded above, because the marginal impact of any one sample on $u_i$ is largest when no agent has yet taken a sample.

First, by direct calculation, we have that for any non-integral $\theta_j$,
\begin{align*}
    \frac{\partial u_i(\btheta)}{\partial \theta_j} 
    =& -\frac 12 \frac{\partial \sum_{x\in\cX} q_{ix} \prod_{l=1}^k \EE{(1-q_{lx})^{m_l}}}{\partial \theta_j}\\
    =& -\frac 12 \frac{\partial \sum_{x\in\cX} q_{ix} \prod_{l\neq j} \EE{(1-q_{lx})^{m_l}}\left((\theta_j -\floor{\theta_j})(1-q_{jx})^{\floor{\theta_j}+1} +(1+\floor{\theta_j}-\theta_j)(1-q_{jx})^{\floor{\theta_j}}\right)}{\partial \theta_j}\\
    =& -\frac 12 \sum_{x\in\cX} q_{ix} \prod_{l\neq j} \EE{(1-q_{lx})^{m_l}}\left((1-q_{jx})^{\floor{\theta_j}+1}-(1-q_{jx})^{\floor{\theta_j}}\right)\\
    =& u_i({\floor{\theta_j}+1},\btheta_{-j}) - u_i({\floor{\theta_j}},\btheta_{-j}) \,.
\end{align*}
For integral-value $\theta_j$, when we increase $\theta_j$ by a small amount $\epsilon\in (0,1)$, $\alpha=\floor{\theta_j+\epsilon} = \floor{\theta_j}$ does not change. Then we have
\begin{align*}
    \frac{\partial_+ u_i(\btheta)}{\partial \theta_j}=& -\frac 12 \frac{\partial_+ \sum_{x\in\cX} q_{ix} \prod_{l\neq j} \EE{(1-q_{lx})^{m_l}}\left((\theta_j -\alpha)(1-q_{jx})^{\alpha+1} +(1+\alpha-\theta_j)(1-q_{jx})^{\alpha}\right)}{\partial \theta_j}\\ 
    =& u_i({\alpha+1},\btheta_{-j}) - u_i(\alpha,\btheta_{-j})\\
    =& u_i({{\theta_j}+1},\btheta_{-j}) - u_i(\btheta) \,.
\end{align*}
When we decrease $\theta_j$ by $\epsilon$,  $\alpha=\floor{\theta_j-\epsilon} = \floor{\theta_j-1}$. Then for all $x\in [\theta_j-1,\theta_j]$, we can represent 
\[u_i(x,\btheta_{-1}) =1-\frac 12 {\sum_{x\in\cX} q_{ix} \prod_{l\neq j} \EE{(1-q_{lx})^{m_l}}\left((x -\alpha)(1-q_{jx})^{\alpha+1} +(1+\alpha-x)(1-q_{jx})^{\alpha}\right)}\,.\]
Thus we have
\begin{align*}
    \frac{\partial_- u_i(\btheta)}{\partial \theta_j}=& -\frac 12 \frac{\partial_- \sum_{x\in\cX} q_{ix} \prod_{l\neq j} \EE{(1-q_{lx})^{m_l}}\left((\theta_j -\alpha)(1-q_{jx})^{\alpha+1} +(1+\alpha-\theta_j)(1-q_{jx})^{\alpha}\right)}{\partial \theta_j}\\ 
    =& u_i({\alpha+1},\btheta_{-j}) - u_i(\alpha,\btheta_{-j})\\
    =& u_i(\btheta)-u_i({{\theta_j}-1},\btheta_{-j})\,.
\end{align*}
Then we argue that for any $\btheta\in \bigtimes_{i=1}^k [0,C_i+1]$, any $t\in \NN\cap [0,C_i+1]$, $u_i(t+1,\btheta_{-j})-u_i(t,\btheta_{-j})= \frac 12 {\sum_{x\in\cX} q_{ix} \prod_{l\neq j} \EE{(1-q_{lx})^{m_l}}q_{jx}(1-q_{jx})^{t}}$ is non-increasing with respect to $t$ and with respect to $\theta_l$ for any $l\neq j$. 

Combining the computing results on sub-gradients and the monotonicity of $u_i(t+1,\btheta_{-j})-u_i(t,\btheta_{-j})$, we know that
\[\frac{\partial u_i(\btheta)}{\partial \theta_i} \geq u_i(C_i+1,\btheta_{-1}) - u_i(C_i,\btheta_{-1})\geq u_i(C_{i}+1,\bC_{-1})-u_i(\bC)\,,\]
and
\[\frac{\partial u_i(\btheta)}{\partial \theta_j}\leq u_i(1,\btheta_{-j}) - u_i(0,\btheta_{-j})\leq u_i(1,\bZero_{-j}) - u_i(0,\bZero_{-j})\leq \frac 12 \sum_{x\in\cX} q_{ix}q_{jx}\leq \frac 12\,.\]

\paragraph{General PAC Learning.} 

In the previous two examples, the utilities are well-behaved over any bounded convex set. However, this might not be true in the general PAC learning case. For example, recall the example in the proof of Theorem~\ref{thm:eqintexist} and let us extend the strategy space $\Theta$ from $\{0,1\}^3$
to $[0,1]^3$ by the randomized rounding method as aforementioned, i.e., 
\[u_i(\btheta) = \frac 12(1+\theta_i+\theta_{i\ominus 1} -\theta_i\theta_{i\ominus 1})\,.\]
Then the utility function is ill-behaved over $[0,1]^3$ since ${\partial u_i(\btheta)}/{\partial \theta_i}  = 0$ when $\theta_{i\ominus 1}=1$. However, it is easy to check that for any $C\in [0,1)$, the utility function is well-behaved over $[0,C]^3$.

\section{Proof of Lemma~\ref{lmm:fixedpt}}\label{app:fixedpt}

\rstfixedpt*

\begin{proof}
The celebrated Brouwer fixed-point theorem states that any continuous function on a compact and convex subset of $\R^k$ has a fixed point. First note that $\bF$ is a well-defined map from $\bigtimes_{i=1}^k [0, \vartheta_i]$ to $\bigtimes_{i=1}^k [0, \vartheta_i]$, which is a convex and compact subset of $\R^k$. All that is left to show is that $\bF$ is a continuous function over $\bigtimes_{i=1}^k [0, \vartheta_i]$.

At a high level, $f$ is continuous because in well-behaved utility functions a small change in other agents' contributions affect the utility of agent $i$ only by a small amount, so a small adjustment to agent $i$'s contribution will be sufficient to meet his constraint. More formally, we show that for any
$\bdelta\in \R^k$ with $\norm{\bdelta}_1\leq 1$, $\lim_{\epsilon \rightarrow 0} \lvert f_i(\btheta)  - f_i(\btheta +\epsilon\bdelta) \rvert =0$.
Define $\btheta' = \btheta +\epsilon\bdelta$, $x = f_i(\btheta)$, and $x'= f_i(\btheta')$.
For every $i$, we have
    \begin{align*}
        u_i\left( x'+\frac{c^i_1\epsilon}{c^i_2},\btheta_{-i}\right) 
        \geq u_i\left( x',\btheta_{-i}\right) + c^i_1\epsilon
        \geq u_i\left( x',\btheta_{-i}\right) + c^i_1\epsilon \|\bdelta_{-i} \|_1
        \geq u_i(x',\btheta_{-i}+\epsilon \bdelta_{-i})
        \geq \mu_i\,,
    \end{align*}
    where the first and third transitions are by the definition of well-behaved functions, and the last transition is by the definition of $\btheta'$ and $x'$.
    This shows that $x\leq x'+\frac{c^i_1\epsilon}{c^i_2}$.
Similarly,
    \begin{align*}
        u_i\left( x+\frac{c^i_1\epsilon}{c^i_2},(\btheta+\epsilon\bdelta)_{-i} \right)     
        \geq u_i(x,(\btheta+\epsilon\bdelta)_{-i})+c^i_1\epsilon
        \geq u_i(x,(\btheta+\epsilon\bdelta)_{-i})+c^i_1\epsilon\norm{-\bdelta_{-i}}_1
        \geq u_i(x,\btheta_{-i})
        \geq \mu_i\,,
    \end{align*}
    which indicates that $x+\frac{c^i_1\epsilon}{c^i_2}\geq x'$. Hence, we have $\abs{x-x'}\leq \frac{c^i_1\epsilon}{c^i_2}$. Therefore, $\bF$ is continuous over $\bigtimes_{i=1}^k [0, \vartheta_i]$.

The proof follows by applying the Brouwer Fixed-Point Theorem.
\end{proof}
\section{More General Construction for Theorem~\ref{thm:eqintexist}}\label{app:general-consruction}

We extend the simple example in Section~\ref{sec:proofofeqexist} into a more general one. 

Consider the domain $\cX = \{0,\ldots,6d-1\}$ for any $d>1$ and the label space $\cY = \{0, 1\}$. We consider agents $\{0,1,2\}$ with distributions $\cD_0,\cD_1,\cD_2$ over $\cX\times \cY$. Similar to the example in Section~\ref{sec:proofofeqexist}, we give a probabilistic construction for  $\cD_0,\cD_1,\cD_2$. Take independent random variables $\bZ_0, \bZ_1, \bZ_2$ that are each uniform over $\{0,1\}^d$. For each $i\in\{0,1,2\}$, distribution $\cD_i$ is a uniform distribution over instance-label pairs $\{((2i+ z_{i,j})d + j, z_{i\ominus 1,j})\}_{j=0}^{d-1}$.
In other words, the marginal distribution of $\cD_i$ is a uniform distribution over $\cX_i=\{x_1,\ldots,x_d\}$ where $x_j$ is equally likely to be $2id + j$ or $(2i+1)d+j$ and independent of other $x_l$ for $l\neq j$. Moreover, the labels of points in distribution $\cD_{i\oplus 1}$ are decided according to the marginal distribution of $\cD_{i}$: if the support of the marginal distribution of $\cD_{i}$ contains $2id+j$, then the points $2(i\oplus 1)d+j$ and $(2(i\oplus 1) +1)d+j$ are both labeled $0$, and if the support of the marginal distribution of $\cD_{i}$ contains $(2i+1)d+j$, then the points $2(i\oplus 1)d+j$ and $(2(i\oplus 1) +1)d+j$ are both labeled $1$.

Consider the optimal classifier conditioned on the event where agent $i$ takes samples $\{((2i + z_{i,j})d+j, z_{i\ominus 1,j})\}_{j\in J_i}$ from $\cD_i$ for all $i$. This reveals $z_{i,j}$ and $z_{i\ominus 1,j}$ for all $j\in J_i$. 
Therefore, the optimal classifier conditioned on this event classifies $(2i + z_{i,j})d+j$ for each $j\in J_i\cup J_{i\ominus 1}$ correctly and misclassifies $(2i + z_{i,j})d+j$ for each $j\notin J_i\cup J_{i\ominus 1}$ with probability $1/2$.

Now we formally define the strategy space and the utility functions that corresponding to this setting. Let $\Theta = \NN^3$ to be the set of strategies in which each agent can take any integral number of samples. Let $u_i(\btheta)$ be the expected accuracy of the optimal classifier given the samples taken at random under $\btheta$. As a consequence of the above analysis,
\begin{align*}
    u_i(\btheta) = 1 -\frac{1}{2d} \sum_{j=0}^{d-1}\left(1-\frac{1}{d}\right)^{\theta_i+\theta_{i\ominus 1}} = 1- \frac{1}{2}\left(1-\frac{1}{d}\right)^{\theta_i+\theta_{i\ominus 1}}\,.
\end{align*}
Then let $\bmu = \mu \bOne$ for any $\mu\in (1/2,1)$ such that $m(\mu):=\ceil{\frac{\log(2(1-\mu))}{\log(1-1/d)}}$ is an odd number. It is easy to find such a $\mu$: arbitrarily pick a $\mu'\in (1/2,1)$; if $m(\mu')$ is odd, let $\mu=\mu'$; otherwise let $\mu=(1-1/d)\mu'+1/d$ such that $m(\mu)=m(\mu')+1$.

Agent $i$'s constraint is satisfied when $\theta_i+\theta_{i\ominus 1}\geq m(\mu)$ and is not satisfied when $\theta_i+\theta_{i\ominus 1}\leq m(\mu)-1$. If $\theta_i+\theta_{i\ominus 1}\geq m(\mu)+1$, agent $i$ can unilaterally decrease her strategy by $1$ and still meet her constraint. Therefore, we have
\[\theta_i +\theta_{i\ominus 1}=m(\mu),\forall i=0,1,2\,.\]
This results in $\theta_0=\theta_1=\theta_2$, which is impossible as $m(\mu)$ is odd and $\theta_i$ is integral for all $i$. Hence, no stable equilibrium over $\Theta = \NN^3$ exists.

\section{Proof of Theorem~\ref{thm:flowergap}}\label{app:flowergap}

\rstflowergap*
\begin{proof}
Consider a family of sets each of size $b = k-1$ demonstrated in Figure~\ref{fig:flower}, where there is one \emph{core agent} that owns $b$ central points and $k-1$ \emph{petal agents} whose sets intersect with that of the core agent.
More formally, let agent $0$ be the core agent whose distribution is uniform over the points $\cX_0 = \{1, \dots, b\}$.
Partition $\cX_0 = \{1, \dots, b\}$ to $\sqrt{b}$ equally sized groups of $\sqrt{b}$ instances $\cX^{1}_0, \dots, \cX^{\sqrt{b}}_0$.
Similarly, partition the $b = k-1$ agents to $\sqrt{b}$ equally sized groups of $\sqrt{b}$ agents $I_{1}, \dots, I_{\sqrt{b}}$.
Each $i \in I_j$ has uniform distribution over the set $\cX_i = \cX^j_0\cup \mathcal{O}_i$, where $\mathcal{O}_i$ is a set of $b-\sqrt{b}$ points that are unique to $i$. The strategy space is $\Theta=\R_+^k$.
\begin{figure}[ht]
    \centering
    \includegraphics[width=0.6\textwidth]{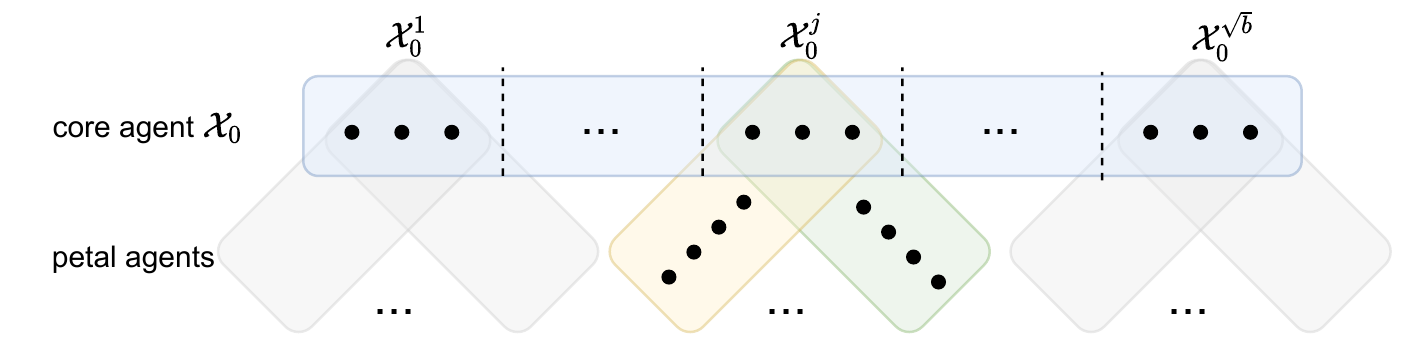}
    \caption{The illustration of the core agent and the petal agents}
    \label{fig:flower}
\end{figure}
Then we consider two learning settings: a) random coverage example and b) linear utility example.

\paragraph{Random Coverage.} Let $m_i\sim \sigma(\theta_i)$ denote the realized integral strategy of agent $i$ for all $i$. For any $I_j$, let $M_j = \sum_{i\in I_j} m_i$ be the total number of samples taken by agents in $I_j$. Then for $i\in I_j$, 
\[
u_i(\btheta)=  1  - \frac{1}{2} \EEs{\bfm}{\frac{1}{\sqrt{b}}\left(1-\frac 1b \right)^{m_0 + M_j} + \left(1-\frac{1}{\sqrt{b}} \right) \left( 1 - \frac 1b \right)^{m_i}}%
\]
and
\[
u_0(\btheta)= 1-\frac{1}{2\sqrt{b}} \EEs{\bfm}{\sum_{j=1}^{\sqrt{b}} \left(1-\frac 1 b\right)^{m_0 + M_j}}\,.
\]
Let $\mu_i=\frac 12 + \frac{1}{2b}$ for all agent $i$.
Note that our choice of $\mu_i$ and distributions implies that the constraint of agent $i$ is met when in expectation at least \emph{one} of the instances in their support is observed by some agent. We use this fact to describe the high level properties of each of the solution concepts.

\underline{The optimal collaborative solution:} Consider the strategy in which the core agent takes 
$O(\sqrt{k})$ samples and all other agents take $0$ samples. This is a feasible solution, because in expectation each $I_j$ receives one of these samples. Therefore, the number of samples in the optimal collaborative solution is at most $O\big( \sqrt{k}\big)$. Specifically, consider the solution in which the core takes $\theta_0 = \ceil{\frac{\ln(1-1/\sqrt{b})}{\ln(1-1/b)}}$ samples and all other agents take $0$ samples.  Let $\btheta^\opt$ denote the socially optimal solution. By direct calculation, it is not hard to check that this is a feasible solution and that $\bOne^\top \btheta^\opt \leq \ceil{\frac{\ln(1-1/\sqrt{b})}{\ln(1-1/b)}}=O(\sqrt{k})$. 

\underline{The Optimal envy-free solution:} By the symmetry of the utility functions for all $i\in I_j$ and for all $j\in \{1, \dots, \sqrt{b}\}$, any envy-free solution must satisfy $\theta_i=\theta$ for some $\theta$ and all $i\in [b]$. This is not hard to check. First, for two petal agents in the same group, i.e., $i,l\in I_j$, and any feasible solution with $\theta_i>\theta_l$, then
\[u_i(\btheta^{(i,l)}) = u_l(\btheta)\geq \mu\,,\]
which indicates that agent $i$ envies agent $l$. Therefore, for any envy-free solution $\theta_i=\theta_l$ for any $i,l\in I_j$. Then for any feasible solution in which any two agents in the same group have the same number of samples, if $\theta_i>\theta_l$ for any $i\in I_j$ and any $l\in I_p$ with $j\neq p$,
\[u_i(\btheta^{(i,l)})\geq u_l(\btheta) \geq \mu\,,\]
which indicates that agent $i$ envies agent $l$.

Furthermore, in any envy-free feasible solution the $0$-th agent's number of sample can be no larger than any other agent. If $\theta_0>\theta$, considering $m_0\sim \sigma(\theta_0)$ and $m\sim \sigma(\theta)$, we have
\[u_0(\btheta^{(0,i)}) = 1-\frac{1}{2}\EEs{\bfm}{\left(1-\frac 1b\right)^{m_0+M_j} +\sum_{p\in [\sqrt{b}]:p\neq j} (1-\frac 1b)^{m_i+M_p}}\geq u_i(\btheta)\geq \mu\,. \]

Let $\btheta^\ef$ represent the optimal envy-free solution. If $\theta_i=\theta>1$ for all $i\in[b]$, we have $\bOne^\top \btheta^\ef =\Omega(k)$. If $\theta\leq 1$, there exists a constant $C>0$ such that for an large enough $b$,
\begin{align}
    u_i(\btheta^{\ef}) =& 1  - \frac 12 \EEs{\bfm}{\frac{1}{\sqrt{b}}\left(1-\frac 1b \right)^{m_0 + \sqrt{b}m} + \left(1-\frac{1}{\sqrt{b}} \right) \left( 1 - \frac 1b \right)^{m}}\nonumber\\
    \leq& 1-\frac{1}{2\sqrt{b}}\left(1-\frac 1b \right)^{\theta_0 + \sqrt{b}\theta} - \frac 12\left(1-\frac{1}{\sqrt{b}} \right) \left( 1 - \frac 1b \right)^{\theta}\label{eq:usingjensen}\\
    \leq& 1-\frac{1}{2\sqrt{b}}\left(1-\frac 1b \right)^{(1 + \sqrt{b})\theta} - \frac 12\left(1-\frac{1}{\sqrt{b}} \right) \left( 1 - \frac 1b \right)^{\theta}\label{eq:usingdef}\\
    \leq & 1-\frac{1}{2\sqrt{b}}e^{-\ln(4) (1 + \sqrt{b})\theta/b} - \frac 12\left(1-\frac{1}{\sqrt{b}} \right) e^{-\ln(4) \theta/b}\nonumber\\
    \leq& 1-\frac{1}{2\sqrt{b}}\left(1-\frac{C(1 + \sqrt{b})\theta}{b}\right) - \frac 12\left(1-\frac{1}{\sqrt{b}} \right) \left(1-\frac{C \theta}{b}\right)\nonumber\\
    \leq& \frac 12 + \frac{3C\theta}{2b}\nonumber\,,
\end{align}
where Eq.~\eqref{eq:usingjensen} adopts Jensen's inequality and Eq.~\eqref{eq:usingdef} uses the property that $\theta_0\leq \theta$. Then since $u_i(\btheta^{\ef})\geq \mu$, we have $\theta\geq \frac{1}{3C}$. Hence, $\bOne^\top \btheta^\ef =\Omega(k)$ and the Price of Fairness is at least $\Omega (\sqrt{k})$.

\underline{The Optimal stable equilibrium:}
First, by the symmetry of the utility functions for all $i\in I_j$ and for all $j\in [\sqrt{b}]$, any stable equilibrium must satisfy $\theta_i=\theta$ for some $\theta$ and all $i\in [b]$. This is not hard to check. For two petal agents $i$ and $l$ in the same group, for any stable feasible solution, if $\theta_i>\theta_l\geq 0$, then $u_i(\btheta)>u_l(\btheta)\geq \mu$, which results in $\theta_i =0$. This is a contradiction. Now for a stable feasible solution in which any two agents in the same group have the same number of samples, if $\theta_i>\theta_l$ for any $i,l$ in different groups, $u_i(\btheta)>u_l(\btheta)\geq \mu$ and thus, $\theta_i=0$. This is a contradiction. Hence, all petal agents have $\theta_i=\theta$ for all $i\in [b]$.

Furthermore, since in any stable equilibrium with $\theta_i=\theta$ for all $i\in [b]$, $u_0(\btheta)>u_i(\btheta)$, agent $0$ must take $0$ samples in any stable equilibrium. Let $\btheta^\eq$ represent the optimal stable equilibrium. Following the similar computation to the case of envy-free solution, if $\theta\leq 1$, we have 
\begin{align*}
    u_i(\btheta^\eq) =& 1  -  \EEs{\bfm}{\frac{1}{2\sqrt{b}}\left(1-\frac 1b \right)^{m_0 + \sqrt{b}m} + \frac 12\left(1-\frac{1}{\sqrt{b}} \right) \left( 1 - \frac 1b \right)^{m}}\nonumber\\
    \leq& 1-\frac{1}{2\sqrt{b}}\left(1-\frac 1b \right)^{\sqrt{b}\theta} - \frac 12\left(1-\frac{1}{\sqrt{b}} \right) \left( 1 - \frac 1b \right)^{\theta}\\
    \leq& \frac 12 + \frac{3C\theta}{2b}\,.
\end{align*}
Therefore, $\theta \in \Omega(1)$, $\bOne^\top \btheta^\eq=\Omega(k)$ and the Price of Stability is at least $\Omega (\sqrt{k})$.

\paragraph{Linear Utilities.} In this flower structure, for any $i\in I_j$,
\[u_i(\btheta) = \theta_i + \frac{1}{\sqrt{b}}(\theta_0+\sum_{l\in I_j:l\neq i}\theta_l) \]
and
\[u_0(\btheta) = \theta_0 + \frac{1}{\sqrt{b}} \sum_{i=1}^{b}\theta_i\,.\]
Let $\bmu = \bOne$. Here the choice of $\bmu$ implies that the constraint of agent $i$ is met when in expectation at least $one$ time, there is an instance being discovered. Similar to the random coverage example, we have the following results.

\underline{The optimal collaborative solution:} There is one feasible solution in which the core agent takes 
$\sqrt{b}$ samples and all other agents take $0$ samples. This is a feasible solution because the core can help every other agent with effort $\frac{1}{\sqrt{b}}$. Let $\btheta^\opt$ denote the socially optimal solution and we have $\bOne^\top \btheta^\opt \leq \sqrt{b} = O(\sqrt{k})$.

\underline{The optimal envy-free solution:} By the symmetry of the utility functions, similar to the random coverage case, any envy-free solution must satisfy $\theta_i=\theta$ for some $\theta$ and all $i\in [b]$. 

Furthermore, in any envy-free feasible solution we must have $\theta_0 \leq \theta$ since $ u_0(\btheta^{(0,i)})\geq u_i(\btheta) \geq 1$. In other words, in any envy-free solution the $0$-th agent's number of sample can be no larger than any other agent, and all other agents take the same number of samples. Let $\btheta^\ef$ denote the optimal envy-free solution. We have
\[1 \leq u_i(\btheta^\ef)\leq \theta + \frac{1}{\sqrt{b}}(\theta+\sum_{l\in I_j:l\neq i}\theta)=2\theta\,,\]
which indicates that $\theta\geq 1/2$. Therefore, $\bOne^\top \btheta^\ef \geq \frac{b}{2}$ and the Price of Fairness is at least $\Omega (\sqrt{k})$.

\underline{The optimal stable equilibrium:}
By the symmetry of the utility functions, similar to the random coverage case, any stable equilibrium must satisfy $\theta_i=\theta$ for some $\theta$ and all $i\in [b]$. Then $u_0(\btheta)=\theta_0+\sqrt{b}\theta$ and $u_i(\btheta) = (2-\frac{1}{\sqrt{b}})\theta + \frac{1}{\sqrt{b}}\theta_0 < u_0(\btheta)$ for $b\geq 2$. Therefore, agent $0$ must take $0$ samples in any stable equilibrium. Then for optimal stable equilibrium $\btheta^\eq$, it is not hard to find that
\[1 \leq u_i(\btheta^\eq)\leq \theta + \frac{\sqrt{b}-1}{\sqrt{b}}\theta\,,\]
which indicates that $\theta \geq \frac{1}{2}$. Therefore, $\bOne^\top \btheta^\eq \geq \frac{b}{2}$ and the Price of Stability is at least $\Omega (\sqrt{k})$.
\end{proof}

\section{Proofs of Theorem~\ref{thm:lineareqopt} and Corollary~\ref{cor:lineareqopt}}\label{app:lineareqopt}

To prove Theorem~\ref{thm:lineareqopt} and Corollary~\ref{cor:lineareqopt}, we first introduce the following three lemmas.

\begin{lemma}\label{lmm:linearpridual}
For any optimal stable equilibrium $\btheta^\eq$ for linear utilities $u_i(\btheta) = W_i^\top \btheta$ and $\mu_i=\mu$ for $i\in[k]$, $\bar{\btheta}^\eq$ is a socially optimal solution for the set of agents $i\in [k]\setminus I_{\btheta^\eq}$, i.e., $\bar{\btheta}^\eq$ is an optimal solution to the following problem.
\begin{equation}\label{eq:linearrm}
\begin{array}{ll}
\min_{\bx} &\bOne^\top \bx \\
\st &\bar{W} \bx\geq \mu \bOne\\
&\bx \geq \bZero\,.
\end{array}
\end{equation}
\end{lemma}

\begin{proof}
The dual problem of Equation~\eqref{eq:linearrm} is 
\begin{equation*}
\begin{array}{ll}
\max_{\by} &\mu \bOne^\top \by \\
\st &\bar{W} \by\leq \bOne\\
&\by \geq \bZero\,,
\end{array}
\end{equation*}
which is equivalent to
\begin{equation}\label{eq:lineardual}
\begin{array}{ll}
\max_\by &\bOne^\top \by \\
\st &\bar{W} \by\leq \mu \bOne\\
&\by \geq \bZero\,.
\end{array}
\end{equation}
Due to the definition of stable equilibrium, for agent $i\in [k]\setminus I_{\btheta^\eq}$, we have $\theta^\eq_i\neq 0$ and thus, $\bar{W} _i^\top \bar{\btheta}^\eq = W_i^\top \btheta^\eq =\mu$. Therefore, $\bar\btheta^\eq$ is a feasible solution to both the primal problem~\eqref{eq:linearrm} and the dual problem~\eqref{eq:lineardual}. This proves that  $\bar\btheta^\eq$ is an optimal solution to Equation~\eqref{eq:linearrm}.
\end{proof}

\begin{lemma}\label{lmm:linearallmu}
   If $\bar\btheta$ is an optimal solution to Equation~\eqref{eq:linearrm}, then $\bar{W}\bar\btheta = \mu\bOne$.
\end{lemma}
\begin{proof}
As proved in Lemma~\ref{lmm:linearpridual}, $\bar{\btheta}^\eq$ is an optimal solution to Equation~\eqref{eq:linearrm} with $\bar W \bar\btheta^\eq = \mu\bOne$. Assume that there exists another optimal solution $\bar\btheta$ such that $\bar W \bar\btheta = \mu\bOne+ \bv$ with $\bv\geq \bZero$. Let $T^* = \bOne^\top\bar\btheta^\eq = \bOne^\top\bar\btheta$ denote the optimal value of Equation~\eqref{eq:linearrm}. Then we have
\[
\bar\btheta^\top \bar W \bar\btheta^\eq=\bar\btheta^\top \mu \bOne = \mu T^*\,,
\]
and 
\[
\bar\btheta^{\eq\top} \bar W \bar\btheta =\bar\btheta^{\eq\top} (\mu \bOne +v)= \mu T^* + \bar\btheta^{\eq\top} \bv\,.
\]
Hence, $\bar\btheta^{\eq\top} \bv=0$. Since $\bar\btheta^{\eq}>\bZero$, then $\bv=\bZero$ and $\bar{W}\bar\btheta = \mu\bOne$.
\end{proof}

Without loss of generality, we let
\[
    W=\begin{bmatrix}\bar W & B\\ B^\top & C\end{bmatrix}\,,
\]
and let $d= k-\abs{I_{\btheta^\eq}}$ denote the dimension of $\bar W$.

\begin{lemma}\label{lmm:linearnull}
   If $\bar\btheta$ is an optimal solution to Equation~\eqref{eq:linearrm}, then we have $B^\top (\bar \btheta^\eq - \bar \btheta) = \bZero$.
\end{lemma}
\begin{proof}
If $\bar W$ is a full-rank matrix, then the optimal solution to Equation~\eqref{eq:linearrm} is unique and thus, $\bar\btheta=\bar\btheta^{\eq}$. 

If $\bar W$ is not a full-rank matrix, we assume that $\bar\btheta \neq \bar\btheta^\eq$. Let $\bv_1,\bv_2,\ldots,\bv_d$ denote the eigenvectors of $\bar W$ with eigenvalues $\lambda_1\geq \lambda_2\geq\ldots\geq \lambda_d$. Since $\bar W$ is not a full-rank matrix, let $d'$ denote the number of zero eigenvalues and we have $\lambda_{d-d'+1} = \ldots \lambda_d =0$. We let $\bb_i$ denote the $i$-th column of $B$ and $c_i = C_{ii}\in[0,1]$.

For any $i\in [k-d]$, let $(\bx,y\be_i)$ with any $\bx\in \R^d,y\in \R$ denote a $k$-dimensional vector with the first $d$ entries being $x$, the $d+i$-th entry being y and all others being $0$s. Since $W$ is PSD, we have
\begin{align*}
    (\bx,y\be_i)^\top W (\bx,y\be_i) = \bx^\top \bar W \bx + 2y\bb_i^\top \bx + c_i y^2 \geq 0\,.
\end{align*}
For any $j=d-d'+1, \ldots, d$, let $\bx = \bv_j$ and $y=-\bb_i^\top \bv_j$, then we have 
\[
    (2-c_i)(\bb_i^\top \bv_j)^2\leq \bv_j^\top \bar W \bv_j = 0\,,
\]
and thus $\bb_i^\top \bv_j=0$ for all $j=d-d'+1, \ldots, d$. By Lemma~\ref{lmm:linearallmu}, we know that $\bar W (\bar \btheta^\eq - \bar \btheta)=\bZero$. Hence $\bar \btheta^\eq - \bar \btheta$ lie in the null space of $\bar W$, i.e., there exists $\balpha\neq \bZero\in \R^{d'}$ such that $\bar \btheta^\eq - \bar \btheta = \sum_{i=1}^{d'}\alpha_i \bv_{d+1-i}$. Then $\bb_i^\top (\bar \btheta^\eq - \bar \btheta) =  \sum_{i=1}^{d'}\alpha_i \bb_i^\top \bv_{d+1-i} = 0$.
\end{proof}

Now we are ready to prove Theorem~\ref{thm:lineareqopt}.
\rstthmlinearopt*
\begin{proof}
  Lemma~\ref{lmm:linearpridual} proves the first part of the theorem. For the second part of the theorem, we prove it by using Lemma~\ref{lmm:linearnull}. For $i\in I_{\btheta^\eq}$, $W_i^\top \tilde{\btheta} = \bar W_i \bar\btheta = \mu$. For $i\in [k]\setminus I_{\btheta^\eq}$, by Lemma~\ref{lmm:linearnull} we have $W_i^\top \tilde{\btheta} = \bb_i^\top \bar\btheta = \bb_i^\top \bar\btheta^\eq = W_i^\top \btheta^\eq \geq \mu$. Therefore, $\tilde{\btheta}$ is a stable equilibrium. Combined with that $\bOne^\top \tilde{\btheta} =\bOne^\top \bar{\btheta} =\bOne^\top \bar{\btheta}^\eq =\bOne^\top {\btheta}^\eq$,  $\tilde{\btheta}$ is an optimal stable equilibrium for agents $[k]$.
\end{proof}

\rstcrllinearopt*
Corollary~\ref{cor:lineareqopt} is a direct result of Theorem~\ref{thm:lineareqopt}.

\section{Proof of Theorem~\ref{thm:lineareqef}}\label{app:lineareqef}
\rstlineareqef*
\begin{proof}
Note that only agents with non-zero number of samples can envy others.
Assume on the contrary that there is agent $i$ with  $\theta^\eq_i>0$ that envies another agent $j$. By the definition of a stable equilibrium, we have that 
$W_i^\top \btheta^\eq =\mu_i$. Let ${\btheta}^{(i,j)}$ represent the strategy with $i$ and $j$'s contributions swapped. Then, 
\[ u_i({\btheta}^{(i,j)})= u_i (\btheta) + (\theta_i - \theta_j)(W_{ij} - W_{ii}) < u_i (\btheta)  = \mu_i,
\]
where the second transition is by $\theta_i> \theta_j$ and $W_{ii} >W_{ij}$. This shows that no agent can have envy in an equilibrium.
\end{proof}
\section{Structure of Equilibria in Random Coverage}\label{app:ncvxef}
In Section~\ref{sec:linearcvx}, we show that the optimal stable equilibrium can be computed by a convex program in the linear case. However, this is not true in random coverage. In the following, we provide an example in which the utility function is non-concave and the the stable feasible set is non-convex. In addition, we provide another example in which the envy-free feasible set is non-convex.

\subsection{Proof of Theorem~\ref{thm:cvrgeqnoncvx}}
\rstthmcvrgeqnoncvx*
\begin{proof}
Let us consider an example where there are $2$ agents and both are with a uniform distribution over the instance space $\cX=\{0,1\}$. Then for any $i\in [2]$, agent $i$'s utility function is
\[u_i(\btheta) = 1-\frac{1}{2}\EEs{\bfm}{\left(\frac{1}{2}\right)^{m_1+m_2}}\,.\]
By direct computation, we have $u_i(\be_1)= u_i(\be_2) = 1-\frac{1}{2}\cdot \frac{1}{2} = \frac{3}{4}$. For $\alpha\in (0,1)$, 
\begin{align*}
    u_i(\alpha \be_1 +(1-\alpha) \be_2) = 1-\frac{1}{2}\left(\alpha\cdot \frac{1}{2} +(1-\alpha) \cdot 1\right)\left((1-\alpha)\cdot \frac{1}{2} + \alpha\cdot 1\right) = \frac 34 - \frac{\alpha(1-\alpha)}{8}\,,
\end{align*}
which is smaller than $\alpha u_i(\be_1) +(1-\alpha) u_i(\be_2)$. Therefore, the utilities in this example are non-concave. 

Let $\mu_i = \frac{3}{4}$ for $i=1,2$. Then, $\be_1$ and $\be_2$ are stable equilibria as no agent has incentive to decrease her number of samples. However, since $\alpha \be_1 +(1-\alpha) \be_2$ is not a feasible solution, the stable feasible set is non-convex.
\end{proof}

\subsection{Proof of Theorem~\ref{thm:cvrgefnoncvx}}
\rstthmcvrgefnoncvx*
\begin{proof}
Now we consider another example showing that the envy-free feasible set is non-convex. Considering the complete graph on $4$ vertices and let each edge correspond to one agent. As illustrated in Figure~\ref{fig:efnoncvx}, we put one point in the middle of every edge and one point on every vertex and let each agent's distribution be a uniform distribution over $\cX_i$, which is the $3$ points on agent $i$'s edge. 
\begin{figure}[t]
    \centering
    \includegraphics[width=0.3\textwidth]{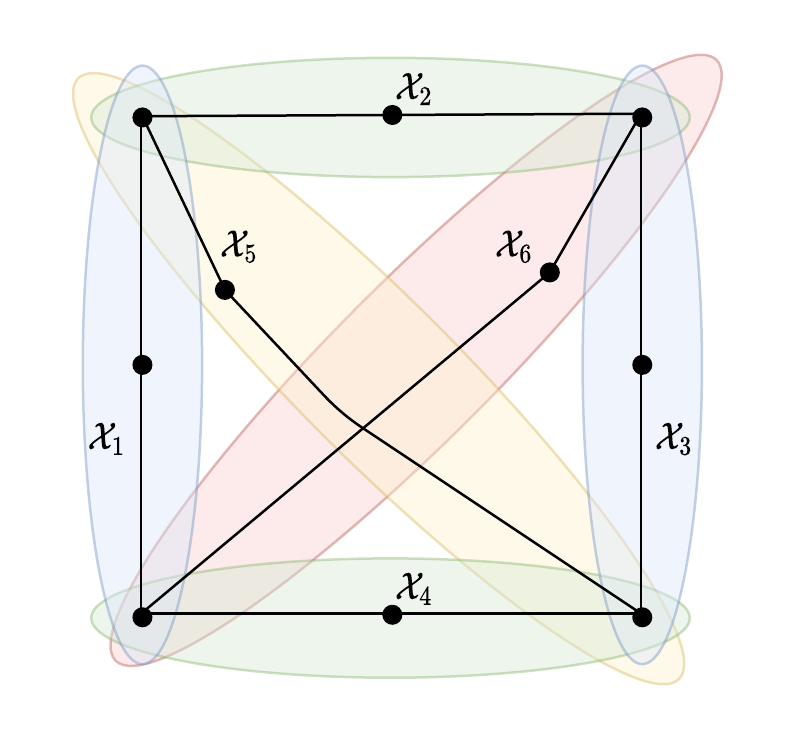}
    \caption{Illustration of the example.}
    \label{fig:efnoncvx}
\end{figure}

Then agent $i$ utility function is
\[
u_i(\btheta) = 1-\frac{1}{6}\EEs{\bfm}{\sum_{x\in \cX_i} (\frac{2}{3})^{n_x}}\,,
\]
where $n_x = \sum_{j:x\in \cX_j} m_j$.
Let $\mu_i=0.6$ for all $i$. Then we consider a solution: pick any perfect matching on this complete graph and then let $\theta_i=1$ if edge $i$ is in this matching and $\theta_i=0$ otherwise. Such a $\btheta$ is an envy-free solution. In this solution, for agent $i$ with $\theta_i=1$, any point $x\in\cX_i$ has $n_x=1$ and the utility is 
\[u_i(\btheta) = 1-\frac{1}{6}\left(3\cdot \frac{2}{3}\right)\geq 0.6\,;\]
for agent $i$ with $\theta_i=0$, two points in $\cX_i$ has $n_x=1$ and one point (in the middle of the edge) has $n_x=0$, and the utility is
\[u_i(\btheta) = 1-\frac 16 \left(2\cdot\frac{2}{3} + 1\right)=\frac{11}{18}\geq 0.6\,.\]
If $\theta_i=1$ and agent $i$ envies another agent $j$ with $\theta_j=0$, agent $i$'s utility after swapping $\theta_i$ and $\theta_j$ is
\[u_i(\btheta^{(i,j)}) = 1-\frac 16\left(\frac{2}{3} + 2\right)=\frac 59<0.6\,.\]
Therefore, this is an envy-free solution.

Then let $\btheta = \be_1 +\be_3$ and $\btheta' = \be_2 +\be_4$. Both are envy-free solutions. Now we show that $\btheta'' = 0.9\btheta+0.1\btheta'$ is not envy-free.
First we show that the agent $2$ meets her constraint in solution  $\btheta''$.
\[u_2(\btheta'') = 1-\frac{1}{6} \left(2\cdot\left(0.09\cdot \left(\frac{2}{3}\right)^2 + 0.82\cdot \frac{2}{3}+0.09\right) + \left(0.1\cdot \frac{2}{3} + 0.9\right)\right)\geq 0.6\,.\]
Now we show that agent $2$ can still meet her constraint after swapping with agent $6$. After swapping $\theta''_2$ and $\theta''_6$, agent $2$'s utility is
\[u_2(\btheta''^{(2,6)}) = 1-\frac{1}{6} \left(\left(0.09\cdot \left(\frac{2}{3}\right)^2 + 0.82\cdot \frac{2}{3}+0.09\right) +1+ \left(0.9\cdot \frac{2}{3} + 0.1\right)\right)\geq 0.6\,.\]
Therefore, $\btheta ''$ is not envy-free and the envy-free feasible set in this example is non-convex.
\end{proof}

\section{Experimental}\label{experimental_section}

\subsection{Dataset}

We use the balanced split of the EMNIST, which is meant to be the broadest split of the EMNIST dataset \citep{DBLP:journals/corr/CohenATS17}. The task consists of classifying English letters and whether they are capitalized or lowercase. Some letters which are similar in their upper and lower case forms, such as C and P, are merged, resulting in just 47 distinct classes. From this dataset, we randomly sample 60,000 points for training and validating the federated learning algorithms. We then take a disjoint sample of an additional 30,000 points to pre-train the model that we will later fine-tune via federation. To select hyperparameters for this model (which we will also use for the federated algorithms), we take the remaining 31,600 points as a validation set. We use top-1 accuracy as the performance metric.

\begin{table}[h]
\centering

\begin{tabular}{|cc|}
\toprule
\textbf{Dataset}                                      & \textbf{Number of Points} \\
\midrule
Potential Training and Validation for Agents & 60,000           \\
Pre-Training                                 & 30,000           \\
Pre-Training Validation                      & 41,600         \\
\bottomrule
\end{tabular}
\end{table}

\subsection{Learning model}
\paragraph{Model} We use a straightforward four-layer neural network with two convolutional layers and two fully-connected layers. We optimize the model with Adam \citep{kingma15} and use Dropout \citep{srivastava2014dropout} for regularization. Architecture details and an implementation can be found via \href{https://github.com/rlphilli/Collaborative-Incentives}{Collaborative-Incentives on Github}. As stated previously, we pre-train the model for $40$ epochs to an accuracy of approximately $55\%$.

\begin{table}[h]
\centering
\begin{tabular}{|ccccc|}
\toprule
\textbf{Algorithm}           & \textbf{Batch Size Per Agent}   & \textbf{Learning Rate} & \textbf{Threshold Accuracy} & \textbf{Local Batches}  \\
\midrule
Individual Learning & 256          & 0.002          & N/A\%   &  N/A\%           \\
FedAvg              & 64           & 0.002          & N/A\%     & 1          \\
MW-FED              & 64 (Average) & 0.002          & 70\%    & 1   \\
\bottomrule        
\end{tabular}
\end{table}

We select hyperparameters using a randomized search on the pre-training validation set. The grid for this search consists of logarithmically-weighted learning rates between $1e-06$ and $1e-02$ and batch sizes of $4$, $8$, $64$, $128$, $256$, and $512$ all together sampled $40$ times. Parameters selected for the individual learning sampling are equivalently translated to the federated learning algorithms.

\begin{algorithm}[H]\caption{\algfont{FedAvg} (simplified to sample all populations each iteration) Let $\eta$ be the learning rate, $m$ be the minibatch size, $B$ be the number of local batches, $k$ be the number of clients, $X_i$ be the set of points for agent $i$, and $\ell$ the loss function }\label{alg:FED}
\begin{algorithmic}[1]
\STATE initialize server weights $\beta_{serv}$ and client weights $\beta_0 \dots \beta_k$
\FOR{each round t=1,2 \dots T }
\FOR{each client $i\in k$}
\STATE $\beta_i \leftarrow{\beta_{serv}}$
\FOR{each local batch $j$ from $1, 2, \dots B$}
\STATE \textbf{sample} $m$ points $x$ from $X_i$
\STATE $\beta_i \leftarrow \beta_i - \eta \nabla \ell \left(\beta_i ; x\right)$
\ENDFOR
\ENDFOR
\STATE $\beta_{serv} \leftarrow  \frac{1}{B \cdot k} \sum_{i=1}^k \beta_i$
\ENDFOR
\STATE return $\beta_{serv}$
\end{algorithmic}
\end{algorithm}

\begin{algorithm}[H]\caption{\algfont{MW-FED} Let $\eta$ be the learning rate, $m$ be the \textit{average} minibatch size, $B$ be the average number of local batches, $k$ be the number of clients, $c$ be the multiplicative factor, $X_i^{train}$ and $X_i^{val}$ be the sets of training and validation points, respectively, for agent $i$, $\epsilon_i$ the desired maximum loss for agent $i$, and $\ell$ the loss function }\label{alg:MW}
\begin{algorithmic}[1]
\STATE initialize server weights $\beta_{serv}$ and client weights $\beta_0 \dots \beta_k$
\STATE initialize contribution-weights $w_1, w_2, \dots w_k = \frac{1}{k}$
\FOR{each round t=1,2 \dots T }
\FOR{each client $i\in k$}
\STATE $\beta_i \leftarrow{\beta_{serv}}$
\STATE $m_i \leftarrow  m \cdot B \cdot \frac{k\cdot w_i}{\sum w}$
\FOR{each local batch $j$ from $1, 2, \dots \floor{\frac{m_i}{m}}$}
\STATE \textbf{sample} $m$ points $x$ from $X_i^{train}$
\STATE $\beta_i \leftarrow \beta_i - \eta \nabla \ell \left(\beta_i ; x\right)$
\ENDFOR
\ENDFOR
\STATE $\beta_{serv} \leftarrow  \frac{1}{ \sum_{i=1}^{k} \floor{\frac{m_i}{m}}} \cdot  \sum_{i=1}^k \beta_i \cdot \floor{\frac{m_i}{m}}  $
\FOR{each client $i\in k$}
    \IF{$\ell\left(\beta_{serv} ; X_i^{val}\right) \geq \epsilon$}
    \STATE $w_i \leftarrow c \cdot w_i$                                 
    \ENDIF 
\ENDFOR
\ENDFOR
\STATE return $\beta_{serv}$
\end{algorithmic}
\end{algorithm}

\subsection{Encouraging heterogeneity across agent datasets}
To encourage heterogeneity between the different agents, we run a series of sampling trials to determine which training points lead to convergence on a holdout data set most quickly. Specifically, over $10,000$ trials we randomly sample the potential agent training set for 1000 points. Then, we train a newly instantiated instance of our network on this data with a batch size of 16 until it reaches a cross-entropy loss of 0.5. For each trial we record the number of iterations it takes for the model to reach 60\% accuracy. At the end of the trials we find the average number of iterations for trials that each point was involved in. The range of these values is from 235 to 670 batches. The mean is 286 and the standard deviation is 21.6 iterations. We then generate agents using mixtures of samples from the top 10\% and bottom 10\% of difficult examples in terms of time to reach the threshold. 

This is an imperfect proxy for difficulty, but we found it useful for producing observable heterogeneity in our chosen samples. We considered other proxies for data value and uncertainty such as output entropy for a sample on the pre-trained model, but found that, in many cases, these samples did not do as much to create differences in how quickly a model trained.

For the main experiment of this section, we create 4 different mixtures : one distribution of 100\% difficult samples, a mixture of 90\% difficult samples, a mixture of 90\% easier samples, and one distribution of 100\% easy samples. As opposed to individual devices, these mixtures might be considered as four different populations with similar, but not identical, objectives. One-hundred averaged training runs for each of these 4 distributions can be found in Figure \ref{fig:solo}.\footnote{Note that, as the batch size differs, these iteration counts can not be directly compared with other statistics in this section.} This figure also shows that they are, in fact, distinct from one another over many repetitions of their training regimes.

\begin{figure}[ht]
    \centering
 \includegraphics[width=.4\textwidth]{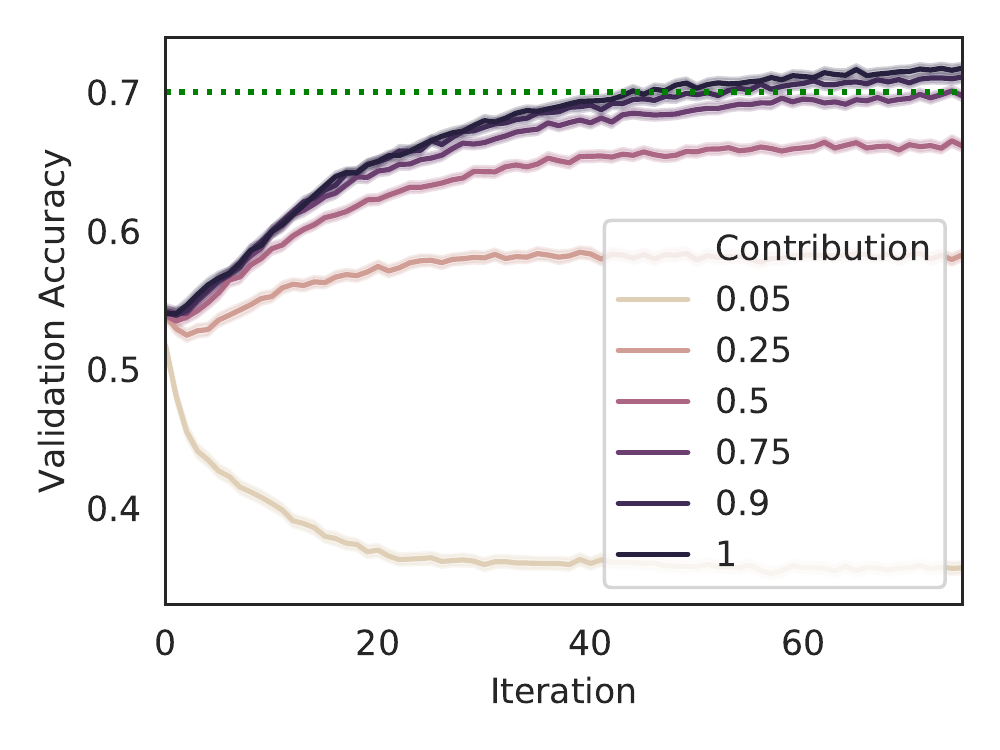}
    \caption{Individual (non-federated) learning averaged over all four agents with different individual contribution levels. At the size of each agent's training dataset (1600), using half or fewer of an agent's unique data points will generally not lead to success.}
    \label{fig:circsolo}
\end{figure}

\paragraph{Non-federated defection is not enough} An important note is that distributions that are often happy while making large defections in the federated settings in Figure \ref{fig:Doublebar} are not generally happy with much less data. Figure \ref{fig:circsolo} shows the averaged learning trajectories over agents who, in the non-federated setting, only use a fraction of their data. In this setting, agents can reduce their contributions by very little if they still hope to be successful. %

\subsection{Connections with algorithms in prior work}
Algorithm \ref{alg:MW} mirrors the multiplicative weights-based solutions that \citet{blum2017collaborative, chen:tight2018, nguyen2018improved} use in the learning-theoretic setting. Specifically, the algorithms in the above prescribe learning in rounds. Each round involves sampling from a weighted mixture of distributions, testing the performance of the learned model on each distribution, and up-weighting those that have not yet reached their performance threshold for the following rounds. %

Section \ref{sec:linearcvx} shows that, in the linear setting, we can use a convex program to find a minimum-cost equilibrium. As previously stated, ensuring there are no $0$-contributors means that we can simply use \ref{eq:LP} to find an equilibrium. Packing LPs such as this are frequently solved using similar multiplicative-weights based strategies \citep{10.2307/3690406, arora_multiplicative_nodate}.

\subsection{Source code}
Code is available \href{https://anonymous.4open.science/r/b3084725-612c-4c6a-8716-61dd0326ec8d/}{in an anonymized repo here.}

\subsection{Computing infrastructure}
The experiments in this work were run using a NVIDIA V100 Tensor Core GPU.

\end{document}